\documentclass[10pt,twocolumn,letterpaper]{article}

\usepackage{cvpr}
\usepackage{times}
\usepackage{epsfig}
\usepackage{graphicx}
\usepackage{amsmath}
\usepackage{amssymb}
\usepackage{subcaption}
\usepackage[pagebackref=true,breaklinks=true,letterpaper=true,colorlinks,bookmarks=false]{hyperref}

\cvprfinalcopy 

\def\cvprPaperID{3460} 

\usepackage{color}

\ifcvprfinal\fi
\begin{document}

\title{ClusterNet: Detecting Small Objects in Large Scenes\\by Exploiting Spatio-Temporal Information}

\author{Rodney LaLonde, Dong Zhang, Mubarak Shah\\
Center for Research in Computer Vision (CRCV), University of Central Florida (UCF)\\
{\tt\small lalonde@knights.ucf.edu, dzhang@cs.ucf.edu, shah@crcv.ucf.edu}
}

\maketitle

\begin{abstract}  \label{abstract}
Object detection in wide area motion imagery (WAMI) has drawn the attention of the computer vision research community for a number of years. WAMI proposes a number of unique challenges including extremely small object sizes, both sparse and densely-packed objects, and extremely large search spaces (large video frames). Nearly all state-of-the-art methods in WAMI object detection report that appearance-based classifiers fail in this challenging data and instead rely almost entirely on motion information in the form of background subtraction or frame-differencing. In this work, we experimentally verify the failure of appearance-based classifiers in WAMI, such as Faster R-CNN and a heatmap-based fully convolutional neural network (CNN), and propose a novel two-stage spatio-temporal CNN which effectively and efficiently combines both appearance and motion information to significantly surpass the state-of-the-art in WAMI object detection. To reduce the large search space, the first stage (ClusterNet) takes in a set of extremely large video frames, combines the motion and appearance information within the convolutional architecture, and proposes regions of objects of interest (ROOBI). These ROOBI can contain from one to clusters of several hundred objects due to the large video frame size and varying object density in WAMI. The second stage (FoveaNet) then estimates the centroid location of all objects in that given ROOBI simultaneously via heatmap estimation. The proposed method exceeds state-of-the-art results on the WPAFB 2009 dataset by $5$-$16\%$ for moving objects and nearly $50\%$ for stopped objects, as well as being the first proposed method in wide area motion imagery to detect completely stationary objects.
\end{abstract}

\section{Introduction}  \label{introduction}

Object detection is a large and active area of research in computer vision. In wide area motion imagery (WAMI), performing object detection has drawn the attention of the computer vision community for a number of years \cite{Nguyen, Reilly, Ruskone, Sommer, Zhao}. Numerous applications exist in both the civilian and military domains. In urban planning, applications include automatic traffic monitoring, driver behavior analysis, and road verification for assisting both scene understanding and land use classification. Civilian and military security is another area to benefit with applications including military reconnaissance, detection of abnormal or dangerous behavior, border protection, and surveillance of restricted areas. With increases in the use and affordability of drones and other unmanned aerial platforms, the desire for building a robust system to detect objects in wide-area and low-resolution aerial videos has developed considerably in recent years.

\begin{figure}[t]
\begin{center}
   \includegraphics[width=0.85\linewidth]{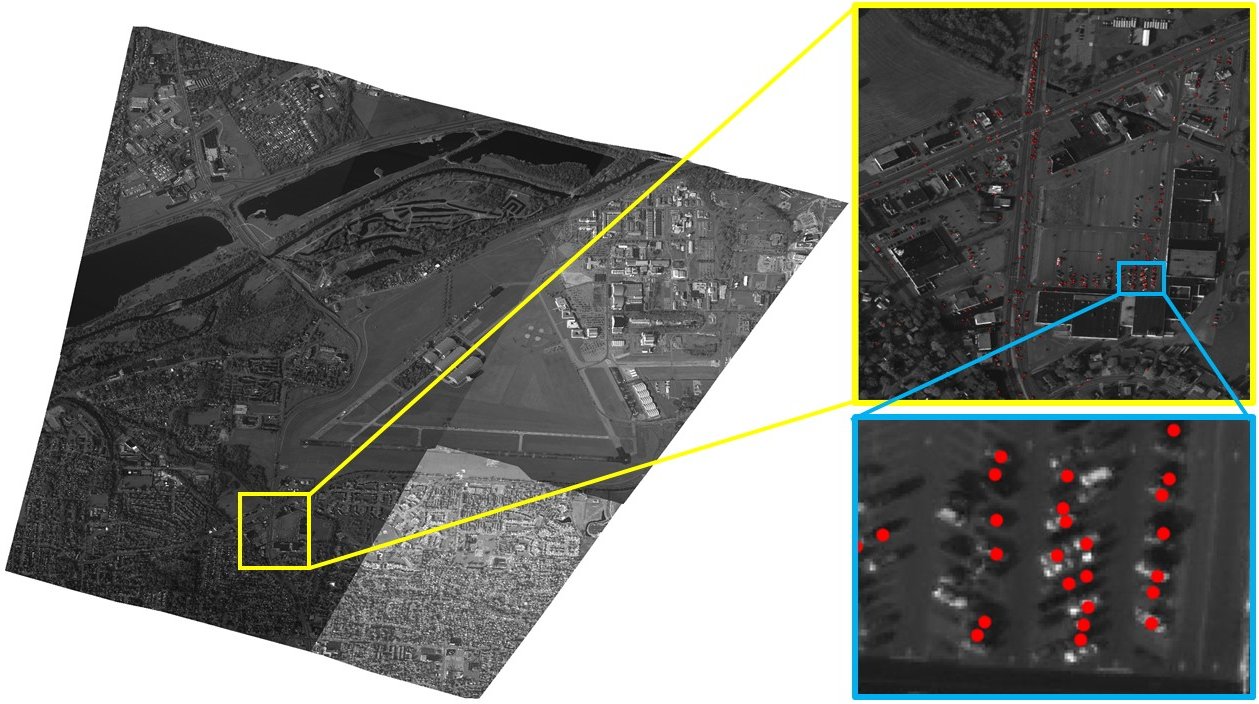}
\end{center}
   \caption{Example WAMI video frame. The yellow-boxed region is enlarged, then a blue-boxed region is further enlarged. Ground truth annotations are marked with red dots.}
\label{fig:WPAFBexample}
\end{figure}

\subsection{Object Detection in WAMI} \label{intro-goal}

\begin{figure}[t]
\begin{center}
   \includegraphics[width=0.8\linewidth]{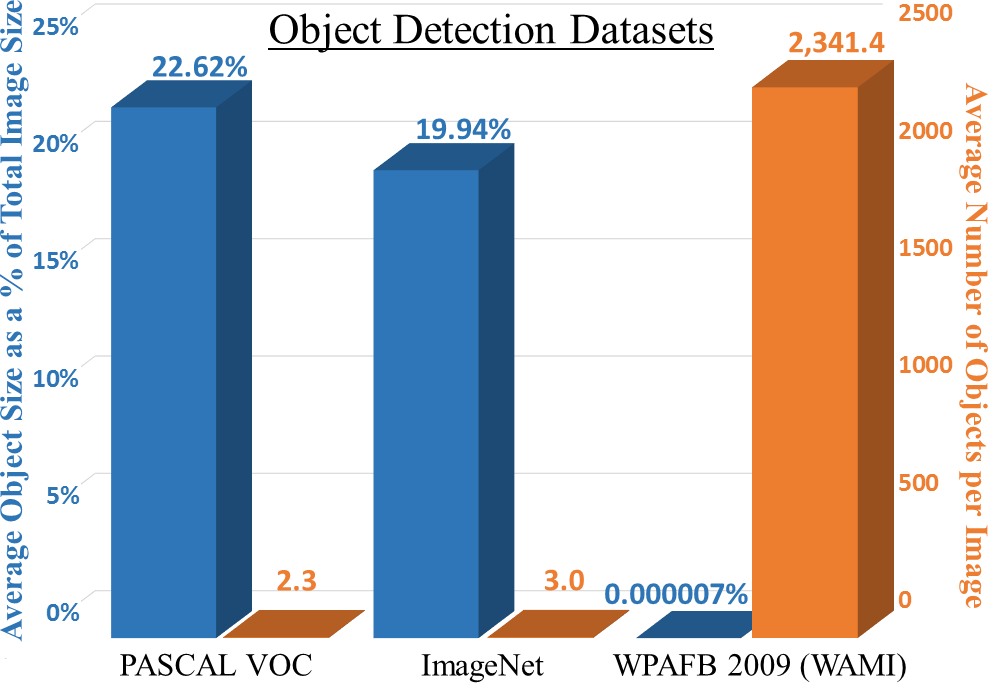}
\end{center}
   \caption{Benchmark datasets in object detection. Two quantities are measured for each dataset: average object size (blue plots, left y-axis) and average number of objects (orange plots, right y-axis).}
\label{fig:DatasetChart}
\end{figure}

The goal of object detection in images or videos is to place a bounding box (\ie the tightest fitting rectangle which contains all elements of an object while containing as few extraneous elements as possible) around all objects in the scene. Object detection in WAMI differs from the typical object detection problem in three major ways: 1) Ground-truth (\ie human-generated) annotations are single $(x,y)$ coordinates placed at the objects' centers, rather than a bounding box. Therefore, scale and orientation invariance must be learned in order to locate objects' centers, but this information cannot be provided during supervised training. 2) In typical object detection datasets, images or video frames most often contain only one to three objects, with no more than 15 objects, while these objects take up a large percentage of the image or video frame \cite{Lin2014}. In WAMI, video frames can contain thousands of small objects, each object accounting for less than $0.000007\%$ of the total pixels in a given frame. Quantitative analysis of this is shown in Fig.~\ref{fig:DatasetChart}. 3) Majority of object detection frameworks deal with images at 256 x 256 up to 500 x 500 pixel resolutions. Video frames in WAMI are significantly larger, typically on the order of several, to hundreds of, megapixels. This creates an extremely large search space, especially given the extremely small typical object size in WAMI being on the order of $9 \times 18$ pixels. An example WAMI video frame with ground-truth annotations is shown in Fig.~\ref{fig:WPAFBexample}.

\subsection{Spatial vs. Temporal Information}  \label{spatial-vs-temporal}

\begin{figure}[t]
\begin{center}
   \includegraphics[width=0.8\linewidth]{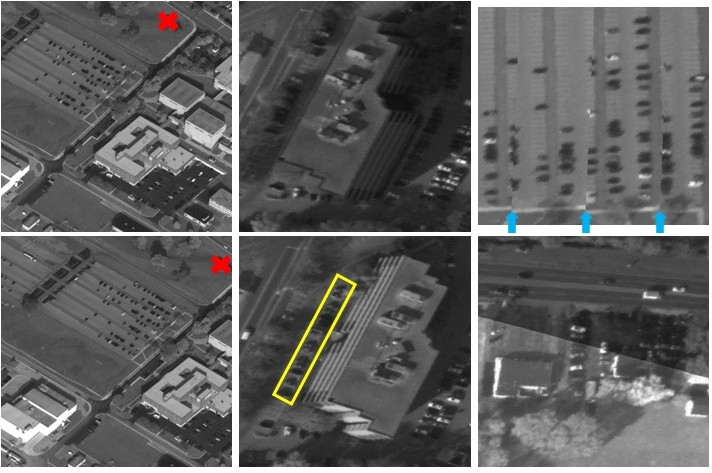}
\end{center}
   \caption{\textbf{Left:} Consecutive video frames showing the large object displacement, illustrated by a red $X$ placed at the same real-world coordinates. \textbf{Center:} Motion parallax effects: Vehicles in the yellow box are occluded at various times. \textbf{Right:} Mosaic seams (blue arrows), camera gain differences, blurred/unclear object boundaries, etc.}
\label{fig:Difficulties}
\end{figure}

For the past several years, object detection has been dominated by detectors relying solely on spatial and appearance information (\eg Faster R-CNN \cite{fasterrcnn}, ResNet \cite{He_2016_CVPR}, YOLO 9000 \cite{yolo9000}). These methods extract low-to-high level spatial and appearance features from images to predict and classify objects. However, it has been stated in numerous recent works \cite{Ren, Saleemi, Sommer, Teutsch} that these appearance- and machine-learning-based methods fail in WAMI due to several unique challenges. 1) Extremely small objects averaging $9 \times 18$ pixels in size. 2) High intra-class variation, ranging from deep black to bright white and from semi-trucks to small cars with the typical vehicle color (\ie silver/gray) exactly matching the background, as well as dramatic changes in camera gain cause significant changes in objects' appearance between consecutive frames. 3) Lacking color and with low resolution, videos are single-channel gray-scale with often blurred/unclear object boundaries. 4) Low frame rates of roughly 1.25 Hz make exploiting temporal information a challenge. Moving objects travel a significant distance between consecutive frames, most often with no overlap to the previous frame. Also, since the aerial recording platform is moving, background objects have significant motion causing strong parallax effects and frame-registration errors, leading to false-positive detections. Moving mosaic seams, where multiple cameras are stitched together to form a single sensor, sweep across the video, leading to even more false positives. Several of these challenges are shown in Fig.~\ref{fig:Difficulties}.

Due to the aforementioned reasons, all state-of-the-art object detection methods in WAMI are motion-based \cite{Prokaj, Sommer, Teutsch}, which use background subtraction or frame differencing to find the objects in the videos. However, as with the appearance-based methods, motion-based approaches suffer from their own costly drawbacks. Frame differencing and background subtraction at their core, rely heavily on the video frame registration. Small errors in frame registration can induce large failures in the final results and attempting to remove false positives is often a big part of these methods. In addition to frame registration, background subtraction requires computing median background images over a large number of frames for the entire video. This combined with the ignorance of appearance information leads to an inefficient use of information across multiple video frames. Yet, the biggest drawback is the complete inability to detect stopped vehicles. All state-of-the-art methods, due to their sole reliance on temporal information, cannot detect vehicles which are not moving relative to the background.

Some recent works \cite{Baccouche, Ji, tubelets, flying, two-stream} have attempted to begin combining spatial and temporal information in various ways for object detection and action recognition. These methods include connecting detections across frames using tracking methods or optical flow, using a sliding window or out-of-the-box detector to perform detection then simply classify this result using some temporal information, as well as combining the outputs of a single frame CNN and optical flow input to a CNN. However, all of these methods rely either on a single-frame detector, which uses no temporal information, or uses a sliding window to check all possible locations in a video frame for object proposals, and thus do not fully exploit temporal information for the task of object detection in video. This is discussed further in Section \ref{related-spatio-temporal}.

\subsection{Contribution} \label{contribution}

The proposed two-stage, spatio-temporal convolutional neural network (CNN) predicts the location of multiple objects simultaneously, without using single-frame detectors or sliding-window classifiers. We show that, consistent with findings in several other works, single-frame detectors fail in this challenging WAMI data, and it is known that sliding window classifiers are terribly inefficient. The novelty of this paper is as follows: 1) Our method effectively utilizes both spatial and temporal information from a set of video frames to locate multiple objects simultaneously in WAMI. 2) This approach removes the need for computing background subtracted images, thus reducing the computational burden and the effect of errors in frame registration. 3) The two-stage network shows the potential to reduce the extremely large search space present in WAMI data with a minimal effect on accuracy. 4) The proposed method is capable of detecting completely stationary vehicles in WAMI, where no other work yet published can do so. 5) The proposed method significantly outperforms the state-of-the-art in WAMI with a $\mathbf{5}$\textbf{-}$\mathbf{16}$\textbf{\% relative improvement in} {$\mathbf{F_1}$} \textbf{score} on moving object detection and a nearly $\mathbf{50}$\textbf{\% relative improvement} for stopping vehicles, while reducing the average error distance of true positive detections from the previous state-of-the-art 5.5 pixels to roughly 2 pixels.

\section{Related Work}  \label{related-work}

\subsection{Frame Differencing \& Background Subtraction} \label{related-WAMI}

As stated in Section \ref{spatial-vs-temporal}, due to the difficulties in WAMI and the reported failures of appearance- and machine-learning-based-methods, all state-of-the-art methods in WAMI are based on either frame-differencing or background subtraction. Both methods require video frames to be registered to a single coordinate system. Reilly \etal \cite{Reilly} detects Haris corners in two frames, computes the SIFT features around those corners, and matches the points using descriptors. A frame-to-frame homography is then fit, using RANSAC or a similar method, and used to warp images to a common reference frame. Frame differencing is the process of computing pixel-wise differences in intensities between consecutive frames. Both two-frame and three-frame differencing methods have been proposed in literature with a number of variations \cite{Keck, Pollard, Saleemi, Sommer, Xiao}. Background subtraction methods focus on obtaining a background model for each frame, then subtract each video frame from its corresponding background model. These methods suffer heavily from false positives introduced by the issues discussed in Section~\ref{spatial-vs-temporal} and cannot detect stationary vehicles. Slowing vehicles also cause a major problem as they are prone to cause split detections in frame differencing \cite{Teutsch} while registration errors and parallax effects are increased in background subtraction models, which use more frames than frame differencing. Sudden and dramatic changes in camera gain cause illumination changes which in-turn cause problems for background modeling and frame differencing methods that require consistent global illumination \cite{Saleemi}. 

\subsection{Region Proposal Networks}

Region proposal networks (RPN), such as Faster R-CNN \cite{fasterrcnn}, which has in some ways become the standard in object detection, have shown the ability to generate object proposals with high accuracy and efficiency. Unfortunately, Faster R-CNN fails in WAMI due to four main reasons. 1) Faster R-CNN acts only on single frames, thus does not exploit the available temporal information, which proves to be extremely important. 2) WAMI video frames are extremely large, thus cannot be sent in their entirety to a Faster R-CNN network on any reasonable number of GPUs. This requires spatially-chipping videos into smaller sections and checking these sections individually, dramatically hurting the computational efficiency benefit supposed to be provided by a RPN. 3) If one changed the RPN stage of Faster R-CNN to extremely downsample the images in the earliest layers in order to fit the large WAMI video frames within GPU memory, object proposals would become impossible. Due to the extremely small object size combined with the areas of high object density means any significant amount of downsampling in the network immediately makes object locations indistinguishable, as they are often separated by only a few pixels or even less. 4) WAMI data is ill-suited for Faster R-CNN as the ground-truth locations are single points, not bounding boxes. We experimentally verify that Faster R-CNN fails in WAMI, even when given the benefit of spatially-chipping the video frames to manageable sizes.

\subsection{Spatio-Temporal CNNs} \label{related-spatio-temporal}

\begin{figure*}[t]
\begin{center}
   \includegraphics[width=0.8\linewidth]{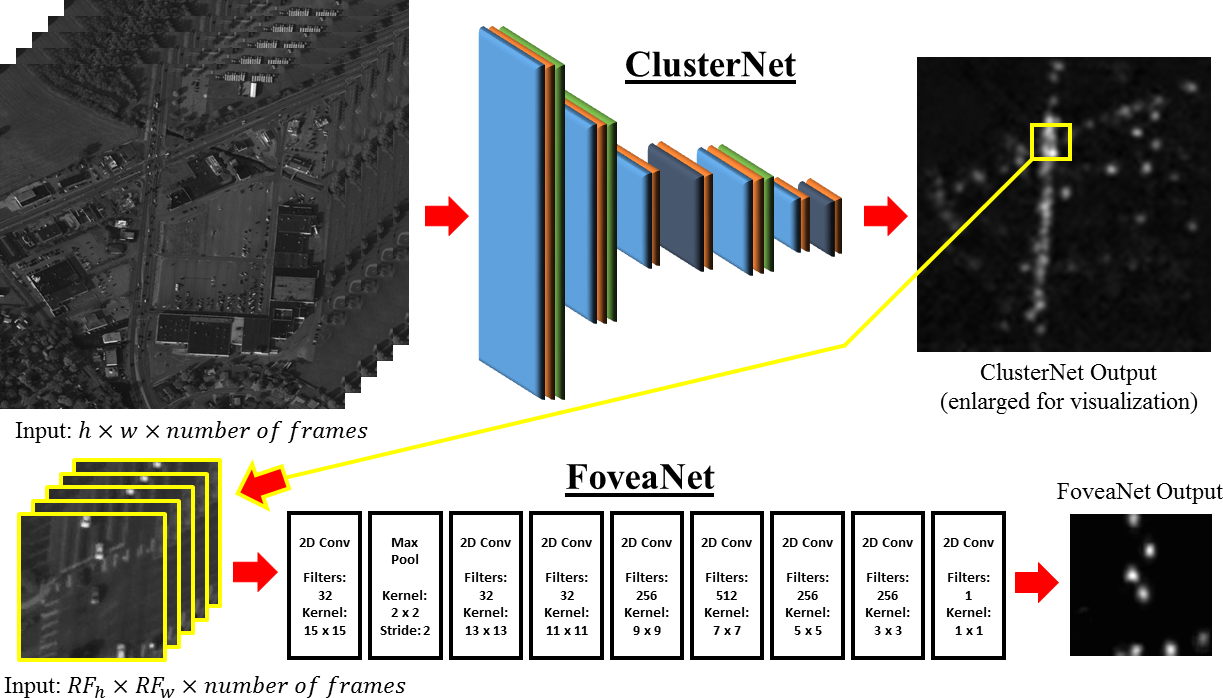}
\end{center}
   \caption{$RF_h$ and $RF_w$ are the height and width of the receptive field of a given set of output neurons. In ClusterNet: light blue and dark blue are $3 \times 3$ and $1 \times 1$ convolutional layers respectively with PReLU activation functions, orange is batch normalization and green in $2 \times 2$ MaxPooling. The $1^{st}$ and $2^{nd}$ convolutional layers have stride $2$. All FoveaNet convolutional layers have ReLU activation functions, and the $6^{th}$ and $7^{th}$ have $50\%$ dropout. Actual results displayed.}
\label{fig:Network}
\end{figure*}

In the past few years, partially due to the enormous success of deep learning methods in a vast array of problems, several works have been proposed for combining spatial and temporal information in various ways within deep learning frameworks. Baccouche \etal \cite{Baccouche} and Ji \etal \cite{Ji} both propose using 3D CNNs for action recognition. Simonyan and Zisserman \cite{two-stream} propose a "two-stream" CNN, one branch receiving individual video frames as input and the other receiving optical flow image stacks where the output of the two streams are combined at the end of the network. Kang \etal \cite{tubelets} proposes several methods to connect object detections in individual frames across time, including using tracking algorithms, optical-flow-guided propagation, and a long short-term memory (LSTM) sub-network. Rozantsev \etal \cite{flying} detects flying drones using sliding-window proposals, input to two CNNs multiple times to align each frame, then performs binary classification of the object or non-object in the sliding window.

Our proposed work differs from all of the above in several key ways. Baccouche \etal and Ji \etal both use stacks of frames as input to a 3D CNN. However, these works do not perform object detection. Both first assume an object of interest is already detected and perfectly centered in each input video frame. To accomplish this, these works use out-of-the-box single-frame human detector algorithms to find the objects of interest in their videos. Our method proposes to solve this object detection problem where single-frame detectors fail, in the challenging WAMI domain. Simonyan and Zisserman keep spatial and temporal information separate during feature extraction, simply combining the extracted features at the end of the network. As stated, single-frame detectors fail in WAMI. Also, due to the extremely large object displacements between consecutive frames, the optical flow stream would likely struggle significantly. The work by Kang \etal also relies on first acquiring single-frame object detections before applying their tracking or LSTM methods. The work by Rozantsev \etal is the only one of these methods which does not rely on single-frame detections, instead opting for a sliding window to first generate its object proposals before using a 3D CNN for classification. However, sliding-window-based methods are extremely inefficient. Our work proposes to generate all object proposals simultaneously using a multi-frame, two-stage CNN for videos in WAMI in a more computationally efficient manner than background subtraction or sliding-windows, effectively combining both spatial and temporal information in a deep-learning-based algorithm.

\section{ClusterNet \& FoveaNet: Two-Stage CNN}

We propose a new region proposal network which combines spatial and temporal information within a deep CNN to propose object locations. Where in Faster R-CNN, each $3 \times 3$ region of the output map of the RPN proposes nine possible objects, our network generalizes this to propose regions of objects of interest (ROOBI) containing varying amounts of objects, from a single object to potentially over $300$ objects, for each $4 \times 4$ region of the output map of the RPN. We then focus the second stage of the network on each proposed ROOBI to predict the location of all object(s) simultaneously for the ROOBI, again combining spatial and temporal information in this network. This two-stage approach is loosely inspired by biological vision where a large field of vision takes in information, then cues, one of the strongest being motion-detection, determine where to focus the much smaller \textit{fovea centralis}. 

\subsection{Region Proposal: Exploiting Motion}

To reduce the extremely large search space in WAMI, several works proposed using road-overlay maps. This dramatically reduces the search area but severely limits to applicability of the method. Road maps must be known in advance and must be fit perfectly to each video frame, in addition to removing the possibility for detecting objects which do not fall on the road. Instead, we proposed a method to learn this search space reduction, without any prior knowledge of road maps. We created a fully-convolutional neural network shown in Fig. \ref{fig:Network} which dramatically downsamples the very large WAMI video frames using convolutional strides and max pooling. To exploit temporal information, rather than sending an individual frame to the CNN, we input consecutive adjoining frames with the frame we want to generate proposals for. These adjoining and central frames are input to a 2D convolutional network. The advantage of using a 2D CNN over a 3D CNN as in \cite{Baccouche, Ji} is the preservation of the temporal relationship between frames. Each frame learns its own convolutional filter set, then these are combined to produce feature maps which maximize information related to the frame we care most about (in our case we chose to train the network to maximize the central frame). Instead of a sliding temporal convolution, our method uses the following equation, 

\begin{equation}
    f_{x,y}^m = \sum_{n=1}^N\Big[\sum_{i=1}^{k_h}\sum_{j=1}^{k_w}V_n(i,j)\times K_n(k_h-i,k_w-j)\Big]+b_m
    \label{eq:conv}
\end{equation}

\noindent where $V_n$ is the $n^{th}$ video frame temporally in the stack and $K_n$ is the convolutional kernel for frame $n$ of size $(k_h, k_w)$, to produce our feature map values $f^m \in \mathbb{R}^M$, where $M$ is the set of feature maps, $n \in N$ is a frame in the set of temporal frames input to the network, and $b_m$ is a learned bias for the feature map $m$. This formulation differs from both the standard 2D single-frame CNN and 3D CNNs by allowing us to choose which frame $n$ we want to maximize via the backpropagation of the Euclidean or cross-entropy loss between the output scoremap and the ground truth heatmap for that desired frame.

All further layers in the network beyond the first perform the task of refining this information to the desired output. As shown by Schwartz-Ziv and Tishby \cite{info-theory}, the amazing success of deep neural networks lie in their "information bottleneck" ability to refine information through the layers guided by backpropagation, reducing the high-entropy input to a low-entropy output. Therefore we chose to provide our temporal information to the network in the earliest layer, providing the maximum possible information at the earliest stage, allowing the remainder of the layers to refine this information to the desired output proposals. 

We formulated the problem in two different ways. In one, we estimate ROOBIs, or object locations in the second stage, via a heatmap-based formulation using the Euclidean loss between the network output and a heatmap created in the following manner, 

\begin{equation}
    H = \sum_{n=1}^N\frac{1}{2\pi\sigma^2}e^{-\frac{(x/2^d)^2+(y/2^d)^2}{2\sigma^2}}
    \label{eq:gauss}
\end{equation}

\noindent where $n \in N$ are single $(x,y)$ ground-truth coordinates, $d$ is the amount of downsampling in the network, and $\sigma$ is the variance of the Gaussian blur fit to each transformed objects location. This gives the loss a smooth gradient to follow for estimating the object/region locations rather than single points in space. $H$ is then clipped at $1$ in order to equally weight regions with single objects and clusters of hundreds of objects. Segmentation maps were created by thresholding the Gaussian heatmaps for our two classes. The segmentation formulation, using a softmax-cross-entropy loss, is used when object locations are mutually exclusive. Therefore, if object locations do not overlap, one could predict the location of a high number of classes of objects using a single output. If locations are not mutually exclusive, the Gaussian heatmap formulation can be employed where each class of object has a corresponding heatmap and the network produces this number of outputs. The results of these experiments show extremely similar results, demonstrating either formulation can be used, given the specific problem, and thus allows our method more flexibility and a wider range of possible applications.

\subsection{FoveaNet: Predicting Object Locations}

\begin{figure}[t]
\begin{center}
   \includegraphics[width=0.8\linewidth]{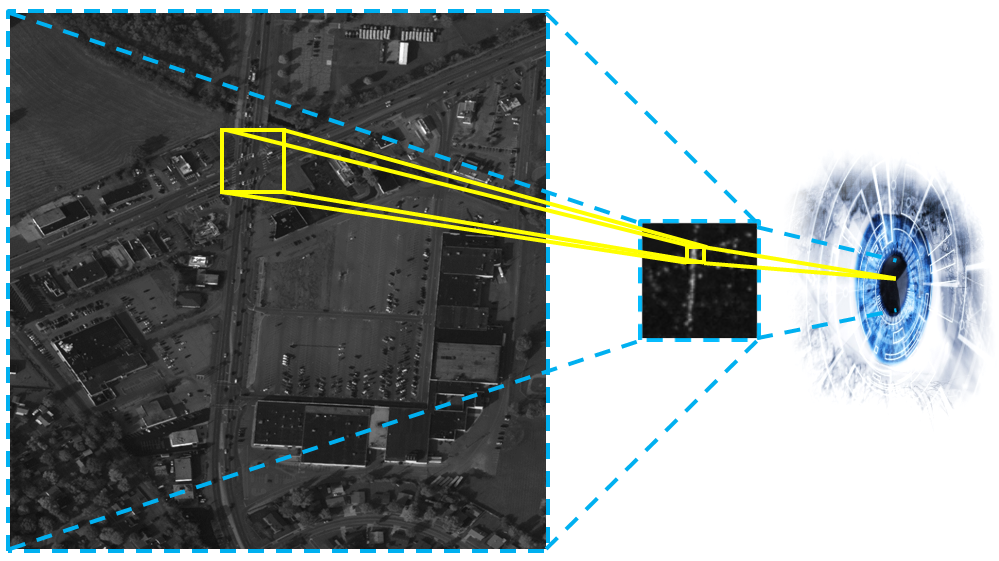}
\end{center}
   \caption{FoveaNet passing over the proposed objects and object clusters from ClusterNet, following up the effective receptive field of high-voting neurons to the initial input.}
\label{fig:foveaDiagram}
\end{figure}

The FoveaNet stage of our two-stage CNN works on the principle of the effective receptive field of neurons in ClusterNet. Each output neuron in the final $1 \times 1$ convolutional layer essentially gives a vote, whether there is a vehicle or cluster of vehicles within that given region or whether there are none. These neurons vote based on the information of the neurons they are connected to in the previous layer which in turn are connected back to the layer before them and so on until the initial input. FoveaNet calculates the region of input information each neuron in the final layer is using to make its final vote. For any ClusterNet output values above a set threshold, this input region, across all input frames, is sent through FoveaNet for high-resolution analysis, as FoveaNet has only a single downsample in the network. The effect is ClusterNet allows us to ignore large regions of the search space while focusing a small high-resolution \textit{fovea centralis} over regions which contain at least one to several hundred vehicles, illustrated in Fig.~\ref{fig:foveaDiagram}. FoveaNet then predicts the location all of vehicles within that region to a high degree of accuracy for the given temporal frame of interest.

Since our FoveaNet input can be much smaller thanks to ClusterNet reducing the search space, we opted to use large kernels within the convolutional layers of FoveaNet, decreasing in size to the final $1 \times 1$ convolutional layer, see Fig \ref{fig:Network}. This was inspired by the recent work by Peng \etal \cite{large-kernel} as well as a large amount of experimentation. For the options of large kernels ascending in size, descending in size, or fixed in size, as well as small kernels, we found the proposed network to consistently perform the best.

\section{Experimental Setup}
\label{experimental-setup}

Experiments were performed on the WPAFB 2009 dataset \cite{WPAFB}. This dataset is the benchmark by which all methods in WAMI compare as it is one of the most varied and challenging, as well as one of the only publicly available with human-annotated vehicle locations. The video is taken from a single sensor, comprised of six slightly-overlapping cameras, covering an area of over $19$ sq. km., at a frame rate of roughly $1.25$ Hz. The average vehicle in these single-channel images make up only approximately $9 \times 18$ out of the over $315$ million pixels per frame, with each pixel corresponding to roughly $1/4$ meter. With almost $2.4$ million vehicle detections spread across only $1,025$ frames of video, there averages out to be well over two thousand vehicles to detect in every frame. 

Frames are registered to compensate for camera motion following the method by Reilly \etal \cite{Reilly} as discussed in Section \ref{related-WAMI}. After registration, eight areas of interest (AOI) were cropped out in accordance to those used is testing other state-of-the-art methods \cite{Basharat, Prokaj, Sommer, Teutsch}, allowing for a proper comparison of results. AOIs $01-04$ are $2278 \times 2278$ pixels, covering different types of surroundings and varying levels of traffic. AOI $34$ is $4260 \times 2604$. AOI $40$ is $3265 \times 2542$. AOI $41$ is $3207 \times 2892$. AOI $42$ is simply a sub-region of AOI $41$ but was included to test our method against the one proposed by Prokaj \etal \cite{Prokaj} on persistent detections where slowing and stopped vehicles were not removed from the ground truth, even though Prokaj \etal uses tracking methods to maintain detections. All other AOIs have any vehicle which moved fewer than 15 pixels ($2/3$ a car length) over the course of $5$ frames removed as to be consistent in testing against other methods for moving object detection. All cropped AOIs are shown with their ground-truth and our results in the supplemental materials.

Data was split into training and testing splits in the following way. For training, only tiles which contain vehicles were included. The splits were as follows: AOIs $02$, $03$, and $34$ were trained on AOIs $40$, $41$, and $42$; AOIs $01$ and $40$ were trained on AOIs $34$, $41$, and $42$; and AOIs $04$, $41$, and $42$ were trained on $34$ and $40$. Both ClusterNet and FoveaNet were trained separately from scratch using Caffe \cite{caffe}. ClusterNet used stochastic gradient descent with Nesterov momentum, a base learning rate of $0.01$, a batch size of $8$, and decreased the learning rate by a factor of $0.1$ upon validation loss plateaus. FoveaNet used Adam \cite{adam} with a base learning rate of $0.00001$ and a batch size of $32$. Training and testing was performed on a single Titan X GPU.

\begin{figure}[t]
\begin{center}
   \includegraphics[width=0.95\linewidth]{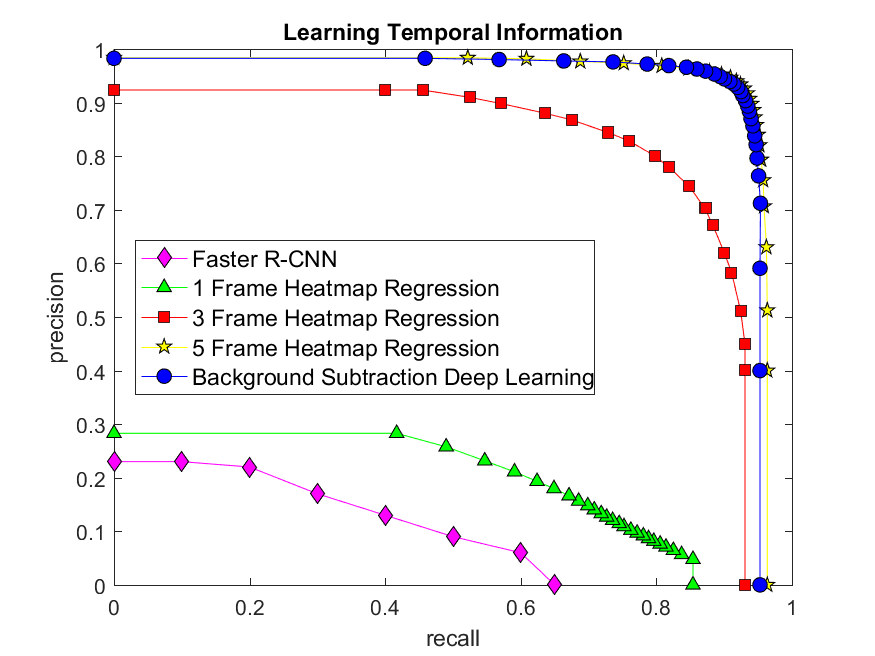}
\end{center}
   \caption{Results on AOI $41$ testing the ability of the deep CNN to learn explicitly or implicitly given temporal information, and its necessity.}
\label{fig:singBSmult}
\end{figure}

To turn the final network output back to single $(x,y)$ coordinates for comparison against the ground-truth, the output is thresholded (either by set levels for creating precision-recall curves, or by Otsu thresholding to find the best threshold level during deployment). Connected components are obtained, weak responses (\ie $< 100$ pixels) are removed, and large responses (\ie $> 900$ pixels; assumed to be merged detections) are split into multiple detections by finding circular centers in a bounding box surrounding that connected component. The centroid of each connected component is considered as a positive detection. It should be noted merged detections are quite rare; completely removing this component saw a $F_1$ score decrease of less than 0.01 across all AOIs. Completely removing small detection removal saw a decrease in $F_1$ score of 0.01 to 0.05 depending on the AOI tested; however, this parameter is quite robust. Values in the range of 60 to 180 pixels show a change of less than 0.01 in F1 score across all AOIs. 

Quantitative results are compared in terms of precision, recall, and $F_1$ measure. To be consistent with literature \cite{Sommer} detections were considered true positives if they fell within $20$ pixels ($5$ meters) of a ground truth coordinate. If multiple detections are within this radius, the closest one is taken and the rest, if they do not have any other ground truth coordinates with $20$ pixels, are marked as false positives. Any detections that are not within $20$ pixels of a ground truth coordinate are also marked as false positives. Ground truth coordinates which have no detections within $20$ pixels are marked as false negatives. 

\section{Results}
\label{results}

\begin{figure}[t]
\begin{center}
   \includegraphics[width=0.95\linewidth]{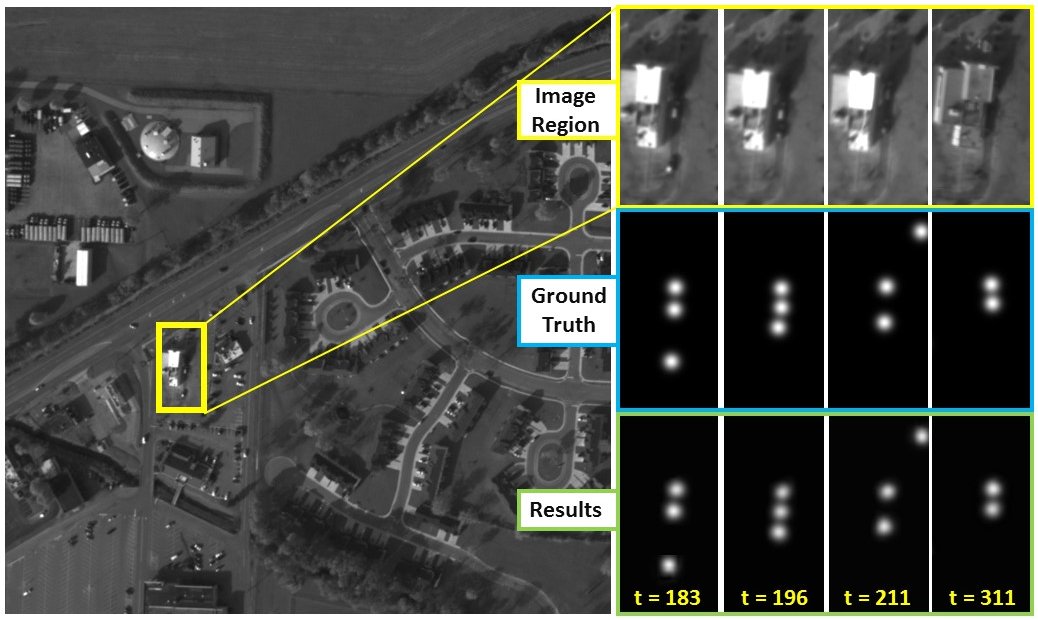}
\end{center}
   \caption{Persistent detection results for AOI 42. The video frame number is marked at the bottom of each column in yellow. \textbf{Top Row:} Highlighted image region at each of the four times. In the first frame shown, the black car in the shadow of the building is nearly invisible to the naked eye. In the last frame shown, due to motion parallax, the white vehicle is nearly completely occluded by the building. \textbf{Middle Row:} Ground-truth heatmap. \textbf{Bottom Row:} Output heatmap without any post-processing.}
\label{fig:AOI42drivethru}
\end{figure}

\subsection{Single-Frame \& Background Subtraction}
\label{single-frame-bs}

To demonstrate the effect of temporal information, we ran three groups of experiments: explicit, implicit, and no temporal information. For explicit, we computed median and background-subtraction images for all frames following the method by Reilly \etal \cite{Reilly}. We then trained and tested FoveaNet using as input two copies of the central video frame combined with the computed background-subtracted image for that frame, each chipped into $128 \times 128$ pixel pieces. This demonstrated our deep network could outperform mere background subtraction through being given both appearance and temporal information. For implicit, we trained and tested our proposed method using three or five frames as input to FoveaNet to demonstrate the networks ability to learn the temporal information directly from the input images, removing the need for computing median and background-subtraction images. For none, we trained and tested FoveaNet using a single frame as input, and conducted experiments using Faster R-CNN. We attempted many configurations of Faster R-CNN with VGG-16 and ResNet-50, pre-trained and trained from scratch, with the proposal sizes tuned to WAMI data split into $256 \times 256$ pixel chips, where the ground-truth bounding boxes were set to $20 \times 20$ pixels centered at each objects location. The highest precision-recall curves for all these experiments, tested on AOI $41$, are shown in Fig.~\ref{fig:singBSmult}.

\begin{figure}[t]
\begin{center}
   \includegraphics[width=0.8\linewidth]{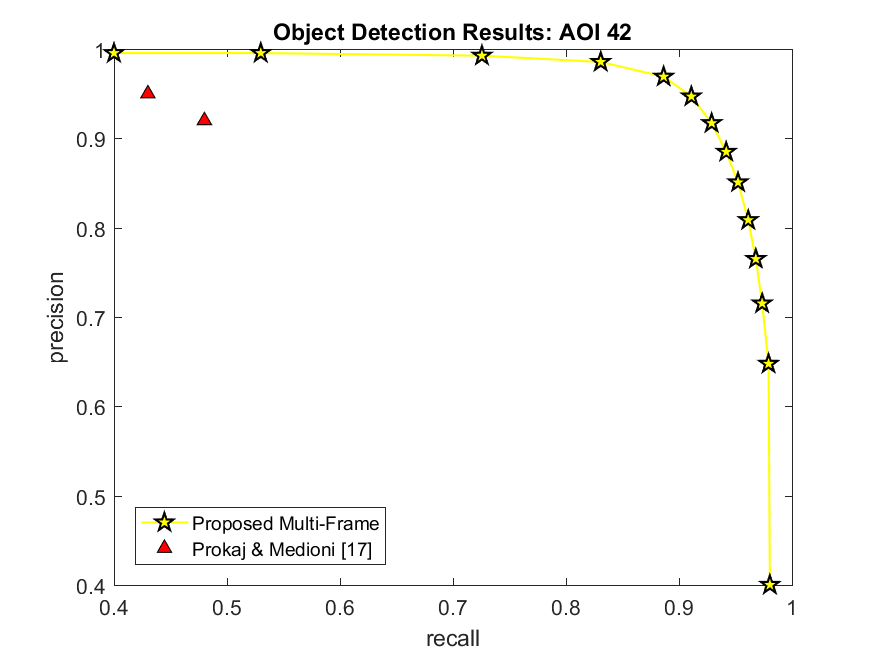}
\end{center}
   \caption{Precision-Recall curve for AOI 42 on persistent detection (\ie no ground-truth coordinates removed).}
\label{fig:AOI42PR}
\end{figure}

\subsection{ClusterNet \& FoveaNet}

\begin{figure*}[t]
\begin{center}
   \includegraphics[width=1\linewidth, trim=0cm 2cm 0cm 0cm, clip=true]{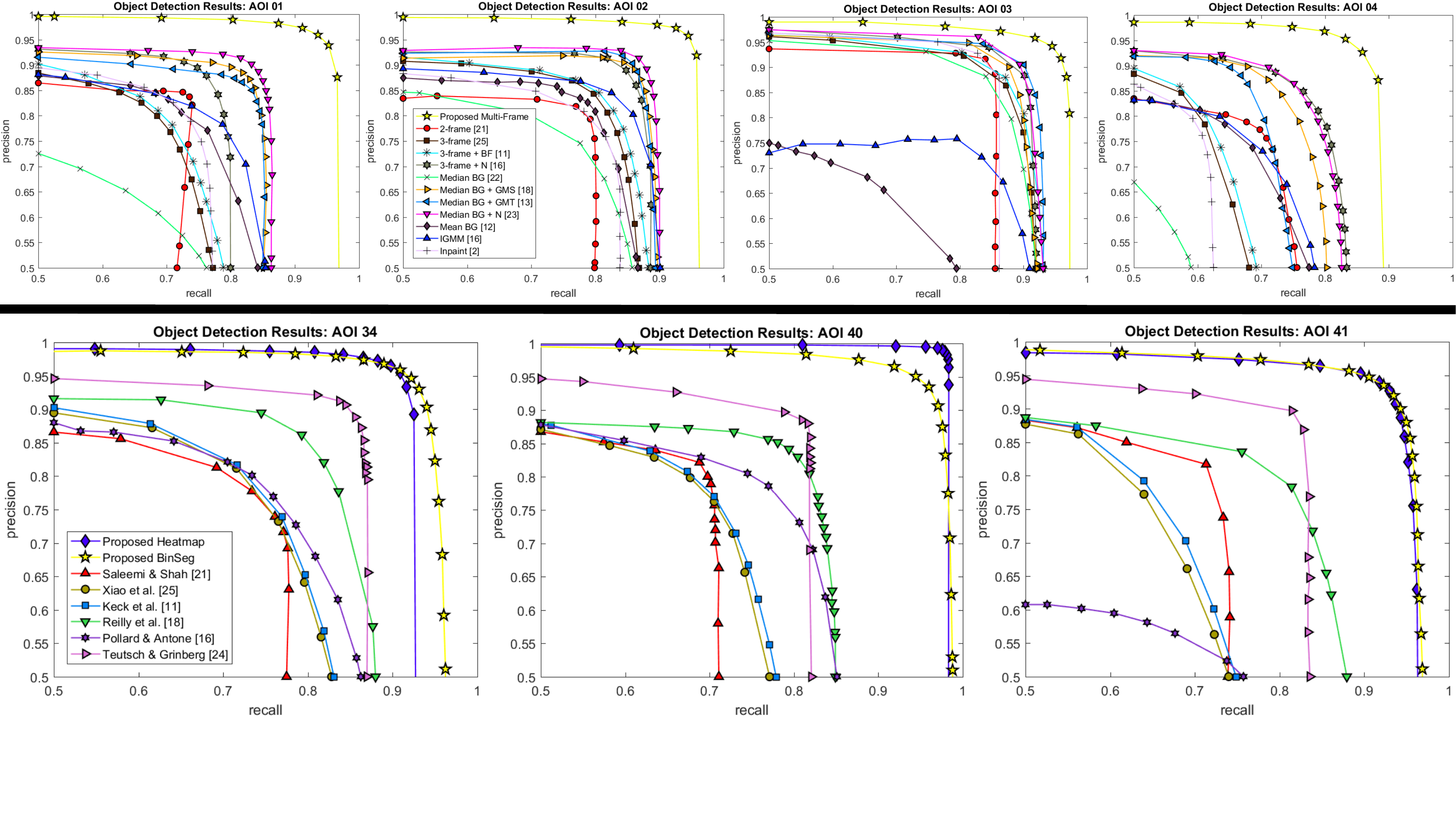}
\end{center}
   \caption{Moving object detection results on seven cropped AOIs with comparisons to $13$ state-of-the-art approaches. If precision-recall values were not reported in the original work, the values reported in \cite{Sommer} and/or \cite{Teutsch} were used.}
\label{fig:PRCurves}
\end{figure*}

The results of our proposed two-stage method, as compared against $13$ different state-of-the-art methods, is shown in Fig.~\ref{fig:PRCurves} and Table \ref{table:Results} across $7$ different AOIs. On AOI $42$, where stationary object are not removed, our results are shown in Fig.~\ref{fig:AOI42PR} and Fig.~\ref{fig:AOI42drivethru}. Our final results measure the computational efficiency improvement and the effect on detection scores provided by ClusterNet. If FoveaNet had to check every single region of a given input, the time to obtain the predicted object $(x,y)$ locations would be roughly $3$ seconds per frame. With that reference, Table \ref{table:Speed-Up} shows the average speed-up from ClusterNet and the associated change in $F_1$ measure averaged across all AOIs.

\begin{table}[t]
\begin{center}
\small
\textbf{Percentage Speed-Up From Using ClusterNet}\par\medskip
\tabcolsep=0.11cm
\begin{tabular}{|c|c|c|c|c|c|}
\hline
Speed-Up & $2-3\%$ & $5-6\%$ & $10-12\%$ & $20\%$ & $30\%$ \\
\hline
$F_1$ Decrease & $0\%$ & $< 1\%$ & $< 3\%$ & $< 5\%$ & $< 8\%$ \\
\hline
\end{tabular}
\end{center}
\caption{Percentage speed-up and $F_1$-measure decrease from using ClusterNet at different threshold levels. Higher thresholds exclude larger portions of the input space, but can negatively impact the $F_1$ score if raised too high.}
\label{table:Speed-Up}
\end{table}

\begin{table*}[t]
\begin{center}
\small
\textbf{Comparison of $F_1$ Scores on Eight Crop and Aligned Sections of the WPAFB 2009 Dataset}\par\medskip
\begin{tabular}{|c|c|c|c|c|c|c|c|c|}
\hline
Method & 01 & 02 & 03 & 04 & 34 & 40 & 41 & 42 \\
\hline
Sommer \etal \cite{Sommer} & 0.866 & 0.890 & 0.900 & 0.804 & x & x & x & x \\
Shi \cite{Shi} & 0.645 & 0.760 & 0.861 & 0.575 & x & x & x & x \\
Liang \etal \cite{Liang} & 0.842 & 0.880 & 0.903 & 0.760 & x & x & x & x \\
Kent \etal \cite{Kent} & 0.767 & 0.807 & 0.668 & 0.711 & x & x & x & x \\
Aeschliman \etal \cite{Aeschliman} & 0.764 & 0.795 & 0.875 & 0.679 & x & x & x & x \\
Pollard \& Antone (3-frame + N) \cite{Pollard} & 0.816 & 0.868 & 0.892 & 0.805 & x & x & x & x \\
Saleemi \& Shah \cite{Saleemi} & 0.783 & 0.793 & 0.876 & 0.733 & 0.755 & 0.749 & 0.762 & x \\
Xiao \etal \cite{Xiao} & 0.738 & 0.820 & 0.868 & 0.687 & 0.761 & 0.733 & 0.700 & x \\
Keck \etal \cite{Keck} & 0.743 & 0.825 & 0.876 & 0.695 & 0.763 & 0.737 & 0.708 & x \\
Reilly \etal \cite{Reilly} & 0.850 & 0.876 & 0.889 & 0.783 & 0.826 & 0.817 & 0.799 & x \\
Pollard \& Antone (IGMM) \cite{Pollard} & 0.785 & 0.835 & 0.776 & 0.716 & 0.766 & 0.778 & 0.616 & x \\
Teutsch \& Grinberg \cite{Teutsch} & x & x & x & x & 0.874 & 0.847 & 0.854 & x \\
Prokaj \& Medioni \cite{Prokaj} & x & x & x & x & x & x & x & 0.631 \\ 
\hline
\textbf{Proposed Multi-Frame} & \textbf{0.947} & \textbf{0.951} & \textbf{0.942} & \textbf{0.887} & \textbf{0.933} & \textbf{0.983} & \textbf{0.928} & \textbf{0.927} \\
\hline
\end{tabular}
\end{center}
\caption{$F_1$ scores of state-of-the-art methods. If $F_1$ values were not reported in the original work, the values reported in \cite{Sommer} and/or \cite{Teutsch} were used. Note that AOI 42 is results on persistent detection (no vehicles removed from ground truth) and is compared with one of the only other persistent detection WAMI methods currently in literature.}
\label{table:Results}
\end{table*}

\section{Conclusion}

We have proposed a novel two-stage convolutional neural network for detecting small objects in large scenes, validated on wide area motion imagery. Our method successfully takes advantage of both appearance and motion cues for detecting the location of single, to hundreds of objects simultaneously. We have shown comparisons with $13$ state-of-the-art methods, and the performance improvements are relatively $5$-$16\%$ on moving objects as measured by $F_1$ score and nearly $50\%$ relative improvement on persistent detections. Additionally, the proposed method's mean distance from ground-truth annotations, averaged over all true positive detections, is roughly $2$ pixels, compared to $5.5$ pixels reported in \cite{Teutsch}. We further demonstrated that the proposed method can detect stopped vehicles, which is not handled by other methods. Removing the computational burden of computing the median and background subtraction images, as well as ClusterNet reducing the search space, are both key contributions to approaching an online method. For future work, one of the final barriers is the removal of frame-alignment computed to remove camera motion.

\section*{Acknowledgement}
The authors would like to acknowledge Lockheed Martin for the funding of this research.

\newpage

\section{Supplemental Materials}

We show more intermediate results in this supplementary material to give the reader a better understanding of our method. In Section 1, we show how ClusterNet and FoveaNet work together to improve the performance; and in Section 2, we show more detailed results on the dataset. Due to the the large size of the figures, we choose to include these qualitative examples of our method and results after the main body of the paper.

\subsection{Two-Stage CNN Visualized With Qualitative Results}

\begin{figure}[t]
  \setlength{\linewidth}{\textwidth}
  \setlength{\hsize}{\textwidth}
  \centering
  \includegraphics[width=5.1in]{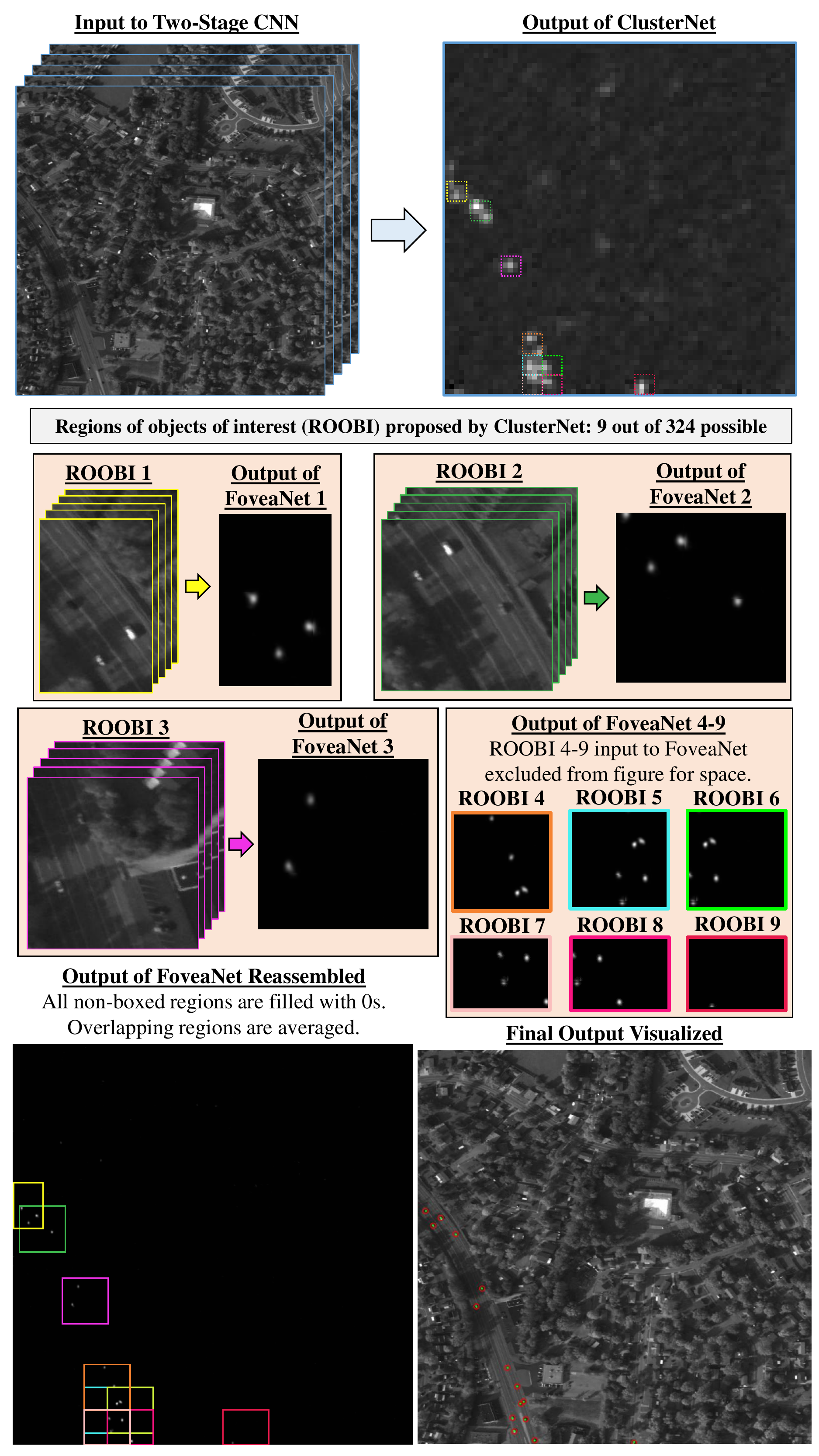}
  \caption{Two-Stage CNN Visualized With Qualitative Results}
  \label{fig:MethodOverview}
\end{figure}

An overview and the performance of different components of the proposed method are shown in Figure~\ref{fig:MethodOverview}. ClusterNet takes as input a set of video frames, containing a very large search space due to the large size of each frame. Each high-scoring $4 \times 4$ region of ClusterNet's output has the associated region of the input space selected, which is selected based on the propagated receptive field of those $16$ output neurons. All low-scoring regions are ignored (set to zero in the output). Working with several to hundreds of megapixel video frames, the frames must be downsampled dramatically early in the network in order to fit within video RAM for deep learning. As a result, localizing individual objects becomes a significant challenge. Each neuron in the output layer of ClusterNet can see anywhere from a single object to none to over $300$, depending on object density. This is best illustrated by the magenta boxed region (corresponding to ROOBI 3), where, even in a sparse area of interest (AOI), a single proposed region of objects of interest (ROOBI) contains two objects separated by a significant distance in the original input space. ROOBIs obtained by ClusterNet are then sent through FoveaNet to simultaneously obtain the final locations of all objects of interest in that region to a high degree of accuracy. The example shown obtains final object locations with perfect precision and recall while needing to check only the $9$ highest-scoring, of the possible $324$, ROOBIs of the output space, saving significant computational time.
 
\cleardoublepage

\cleardoublepage

\subsection{Qualitative Results and ROC Curves}

\begin{figure}[htp]
  \setlength{\linewidth}{\textwidth}
  \setlength{\hsize}{\textwidth}
  \centering
  \caption{\textbf{Left Column Top:} Output of ClusterNet for the given frame shown at right. \textbf{Left Column Bottom:} Receiver operator curves (ROC) to compliment the precision-recall curves in the main paper. \textbf{Right Column:} Final output of the proposed two-stage framework for example frames for AOIs of the WPAFB 2009 dataset. Red Circles are centered on ground truth coordinates. Green dots are the final predicted object locations by the proposed framework. \\[0.5em]}
  \begin{subfigure}[t]{0.95\textwidth}
      \includegraphics[width=2.446in]{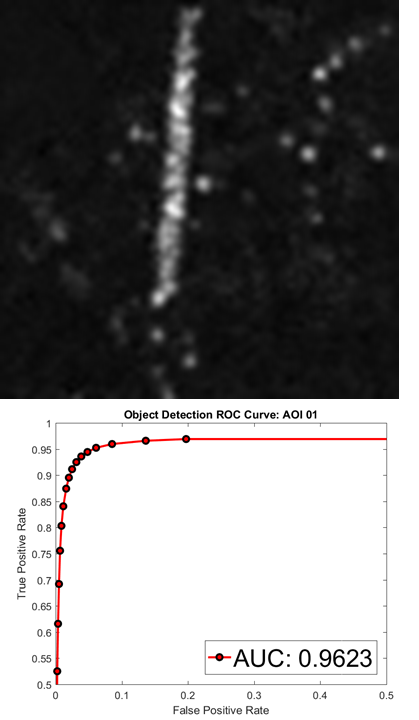}%
      \includegraphics[width=4.2in]{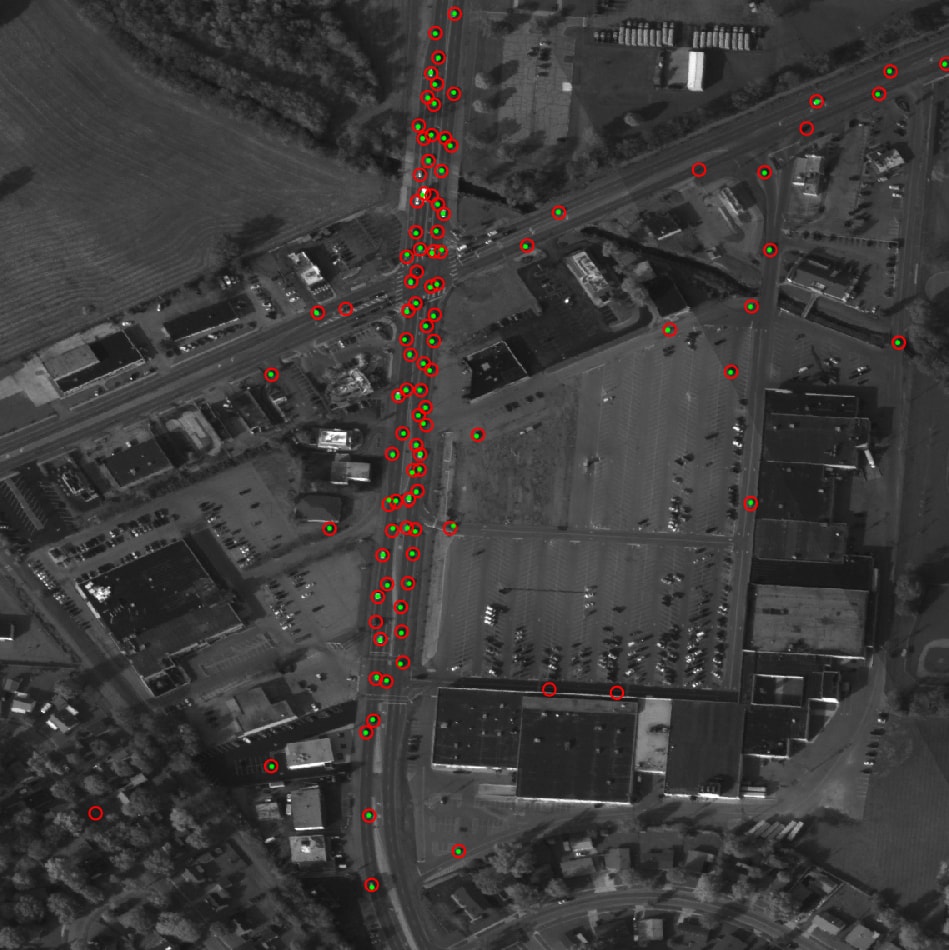}%
      \caption{AOI 01 results using 5-frames and the Gaussian heatmap formulation. ClusterNet output shown at left; FoveaNet output and ground-truth shown at right.}
      \label{fig:AOI01ResultAndROC}
  \end{subfigure}
  \label{fig:Exemplars1}
  \end{figure}
  
\addtocounter{figure}{-1}

\begin{figure*}[t]
\begin{center}
    \begin{subfigure}[t]{0.95\textwidth}
        \addtocounter{subfigure}{1}
        \begin{center}
            \includegraphics[width=2.33in]{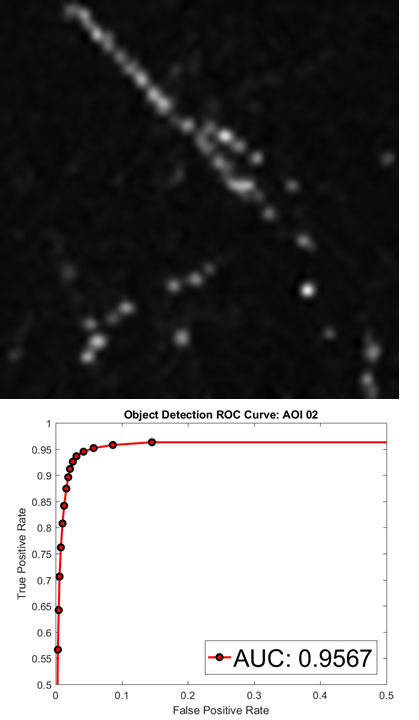}
            \includegraphics[width=4in]{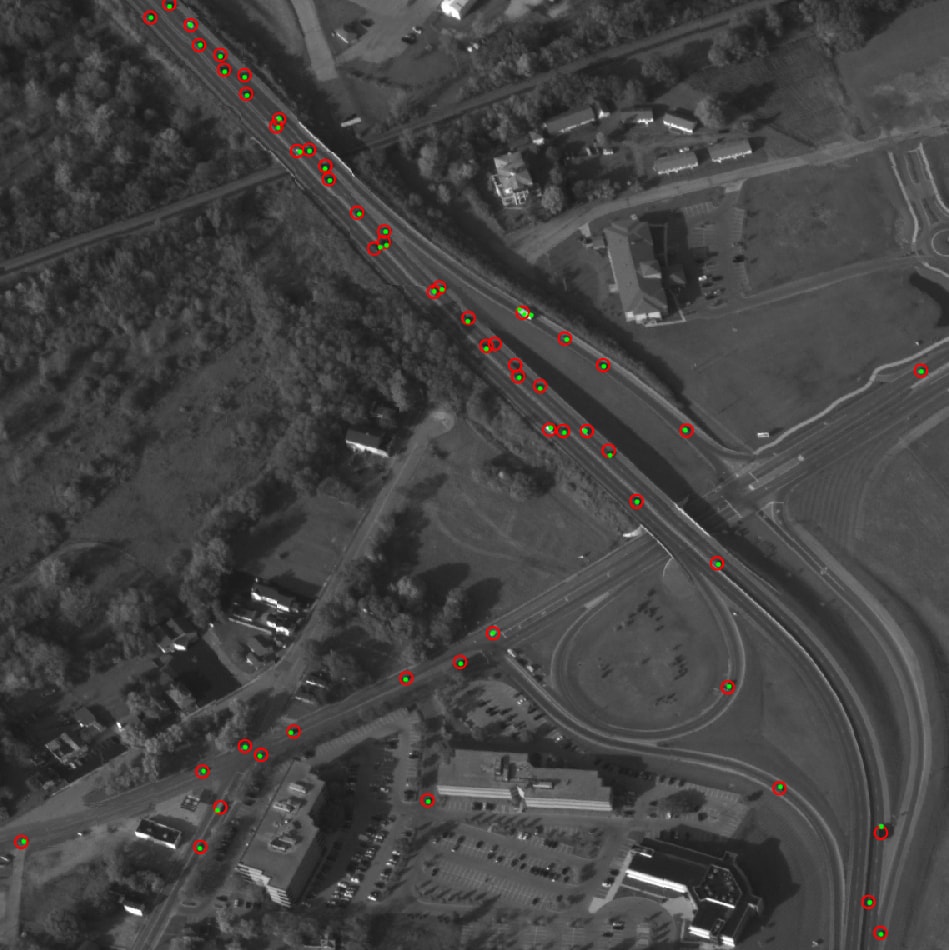}
            \caption{AOI 02 results using 5-frames and the Gaussian heatmap formulation. ClusterNet output shown at left; FoveaNet output and ground-truth shown at right.}
            \label{fig:AOI02ResultAndROC}
        \end{center}
    \end{subfigure}\\[0.5em]
    \begin{subfigure}[t]{0.95\textwidth}
        \begin{center}
            \includegraphics[width=2.33in]{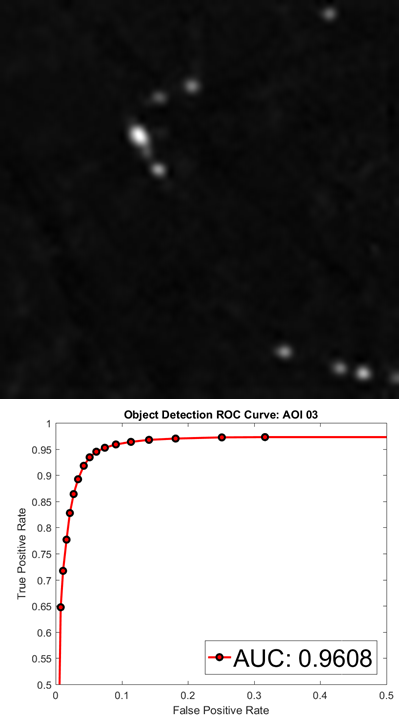}
            \includegraphics[width=4in]{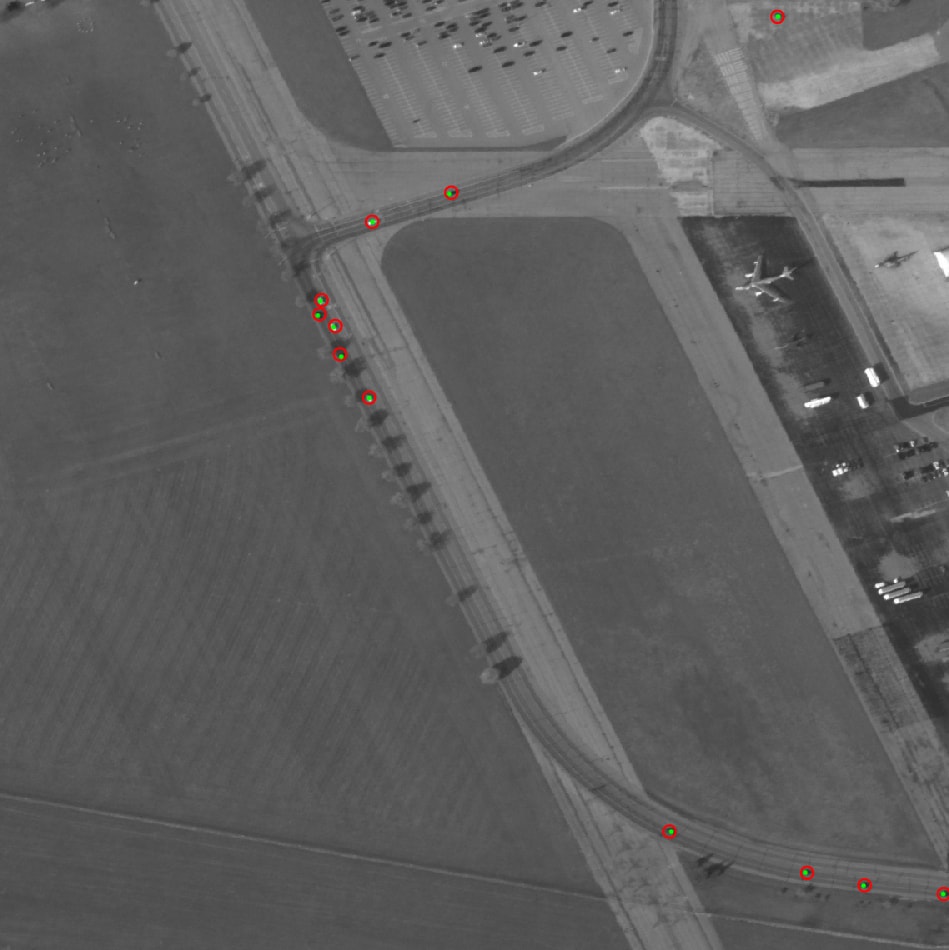}
            \caption{AOI 03 results using 5-frames and the Gaussian heatmap formulation. ClusterNet output shown at left; FoveaNet output and ground-truth shown at right.}
            \label{fig:AOI03ResultAndROC}
        \end{center}
    \end{subfigure}
\end{center}
\label{fig:Exemplars2}
\end{figure*}

\addtocounter{figure}{-1}

\begin{figure*}[t]
\begin{center}
    \begin{subfigure}[t]{0.95\textwidth}
        \addtocounter{subfigure}{3}
        \begin{center}
            \includegraphics[width=2.33in]{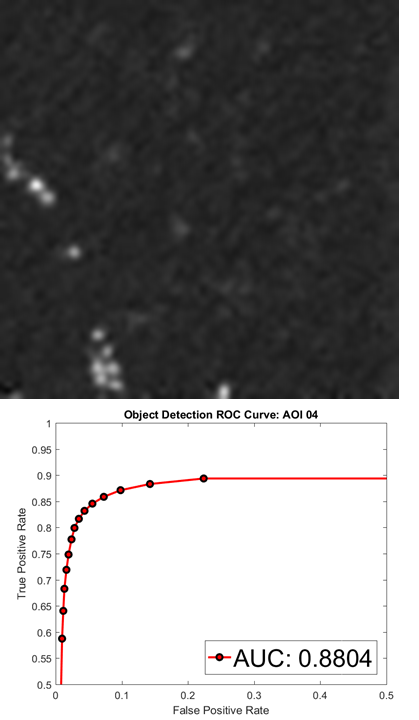}
            \includegraphics[width=4in]{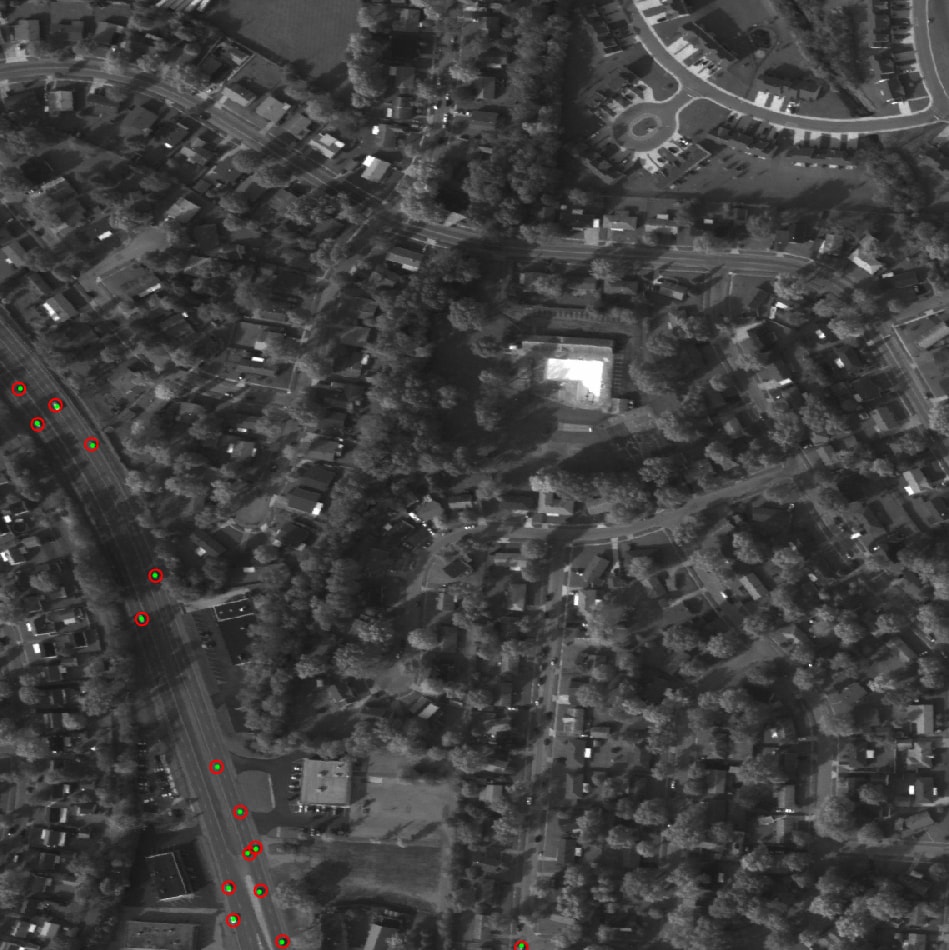}
            \caption{AOI 04 results using 5-frames and the Gaussian heatmap formulation. ClusterNet output shown at left; FoveaNet output and ground-truth shown at right.}
            \label{fig:AOI04ResultAndROC}
        \end{center}
    \end{subfigure}\\[0.5em]
    \begin{subfigure}[t]{0.95\textwidth}
        \begin{center}
            \includegraphics[width=2.33in]{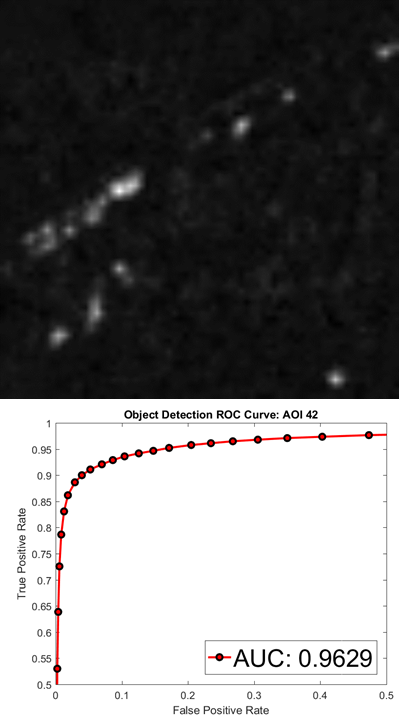}
            \includegraphics[width=4in]{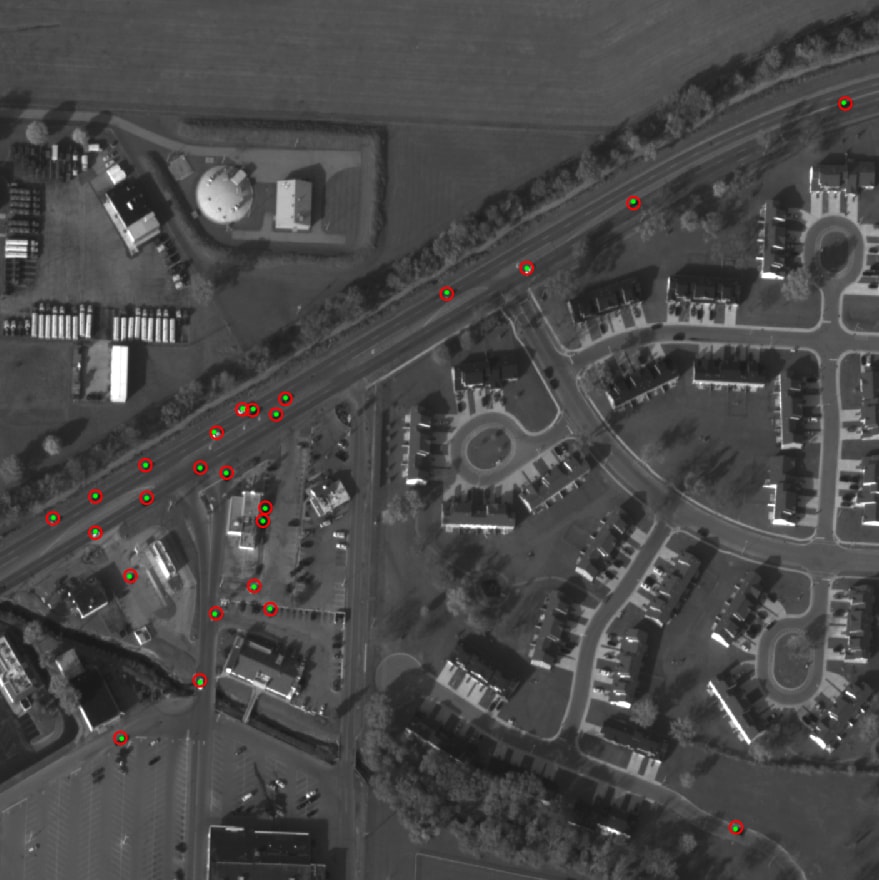}
            \caption{AOI 42 results using 5-frames and the Gaussian heatmap formulation. ClusterNet output shown at left; FoveaNet output and ground-truth shown at right. Note AOI 42 contains all ground-truth coordinates, stopped vehicles are not removed.}
            \label{fig:AOI42ResultAndROC}
        \end{center}
    \end{subfigure}
\end{center}
\label{fig:Exemplars3}
\end{figure*}

\addtocounter{figure}{-1}

\begin{figure*}[t]
\begin{center}
    \begin{subfigure}[t]{0.95\textwidth}
        \addtocounter{subfigure}{5}
        \begin{center}
            \includegraphics[width=2in]{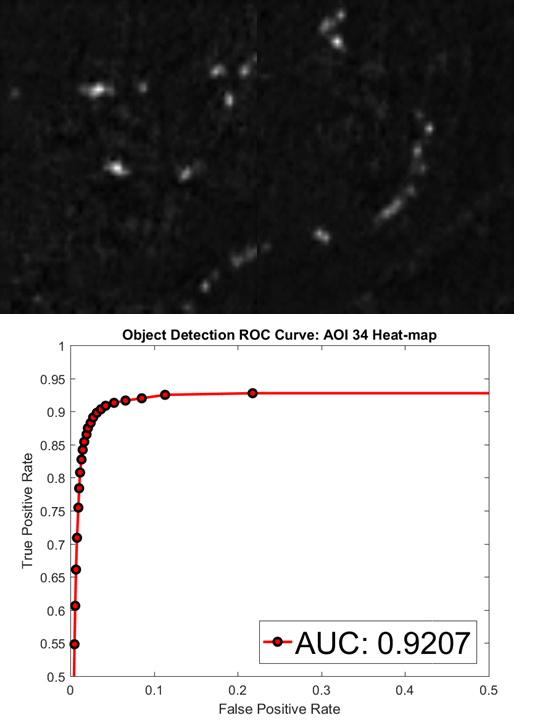}
            \includegraphics[width=4.4in]{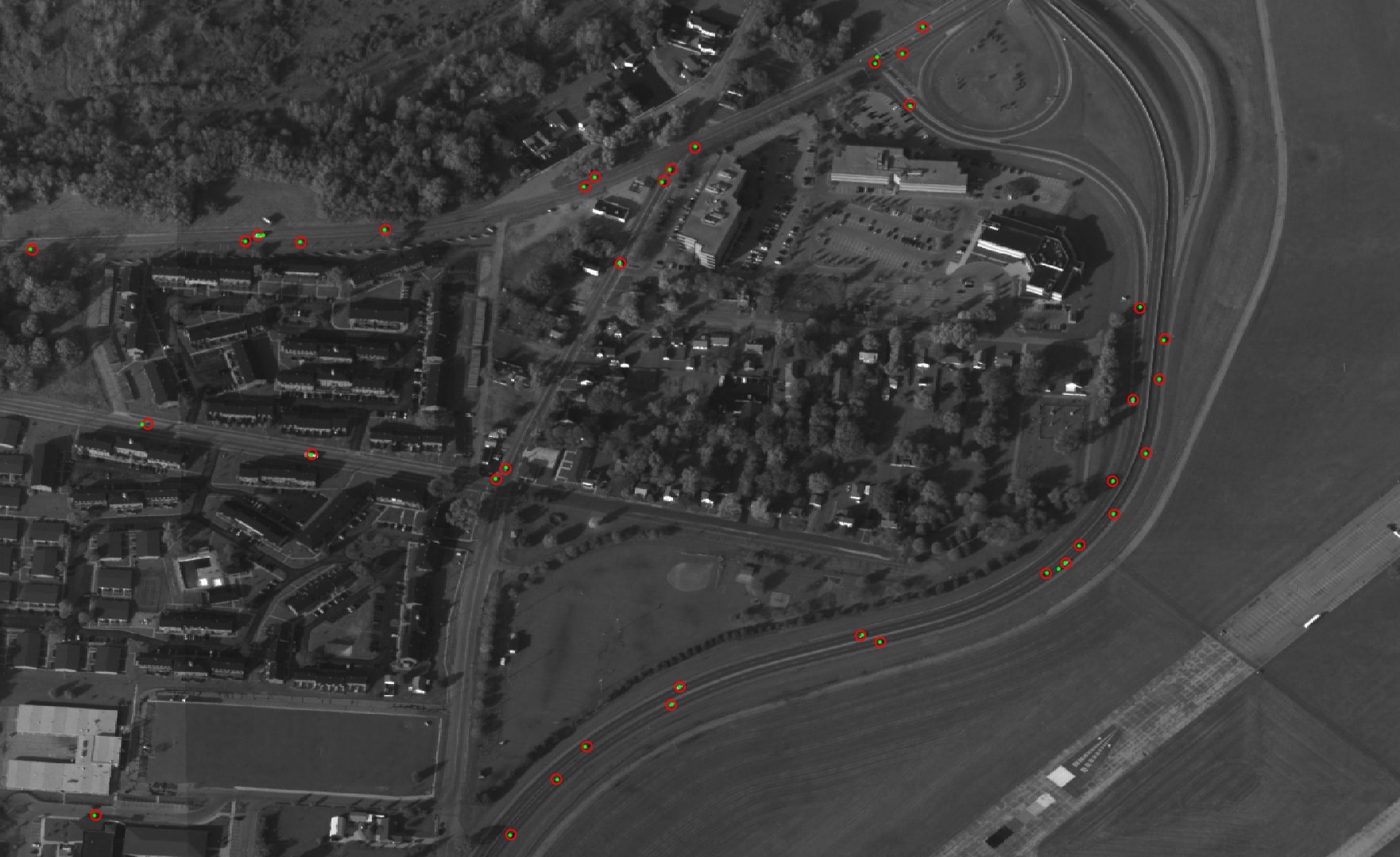}
            \caption{AOI 34 results using 5-frames and the Gaussian heatmap formulation. ClusterNet output shown at left; FoveaNet output and ground-truth shown at right.}
            \label{fig:AOI34ResultAndROC}
        \end{center}
    \end{subfigure}\\[2em]
    \begin{subfigure}[t]{0.95\textwidth}
        \begin{center}
            \includegraphics[width=2in]{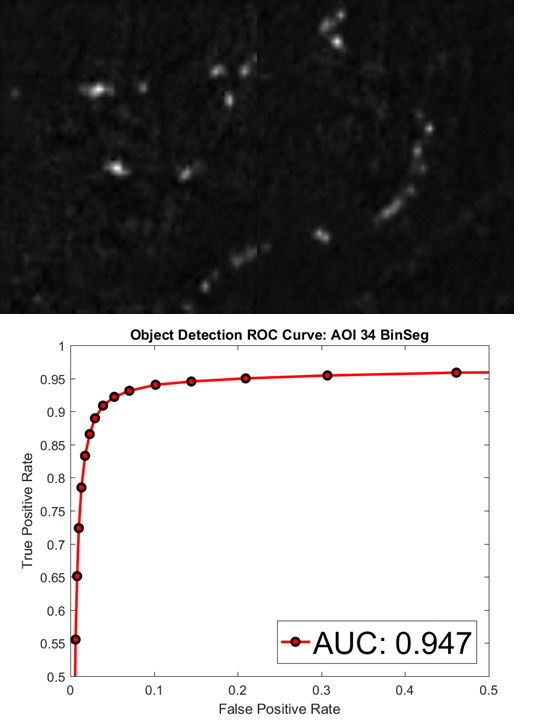}
            \includegraphics[width=4.4in]{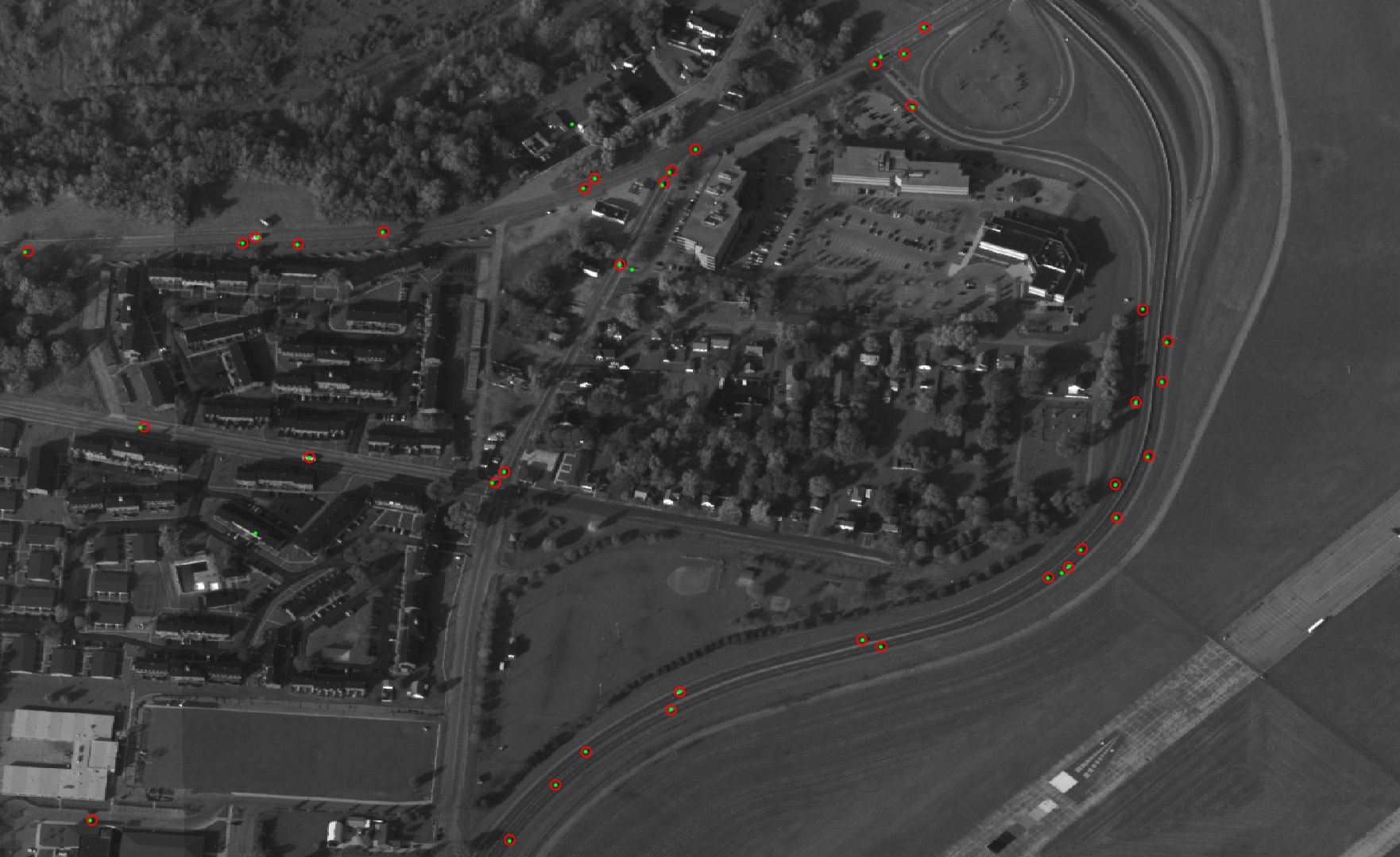}
            \caption{AOI 34 results using 5-frames and the binary segmentation formulation for FoveaNet. ClusterNet output shown at left; FoveaNet output and ground-truth shown at right.}
            \label{fig:AOI34SMResultAndROC}
        \end{center}
    \end{subfigure}
\end{center}
\label{fig:Exemplars4}
\end{figure*}

\addtocounter{figure}{-1}

\begin{figure*}[t]
\begin{center}
    \begin{subfigure}[t]{0.95\textwidth}
        \addtocounter{subfigure}{7}
        \begin{center}
            \includegraphics[width=2.145in]{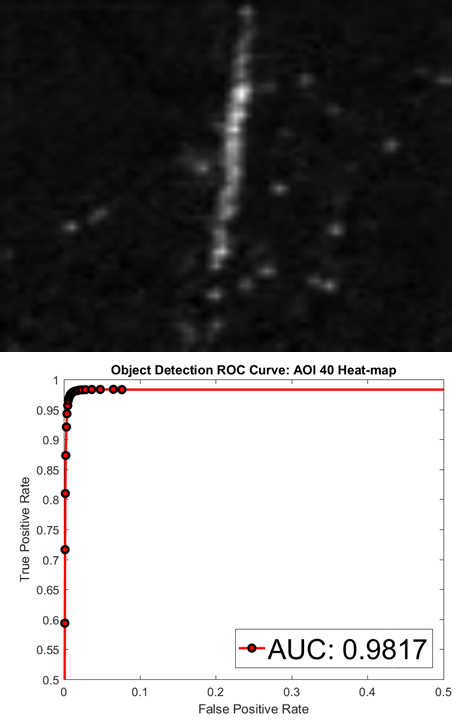}
            \includegraphics[width=4.2in]{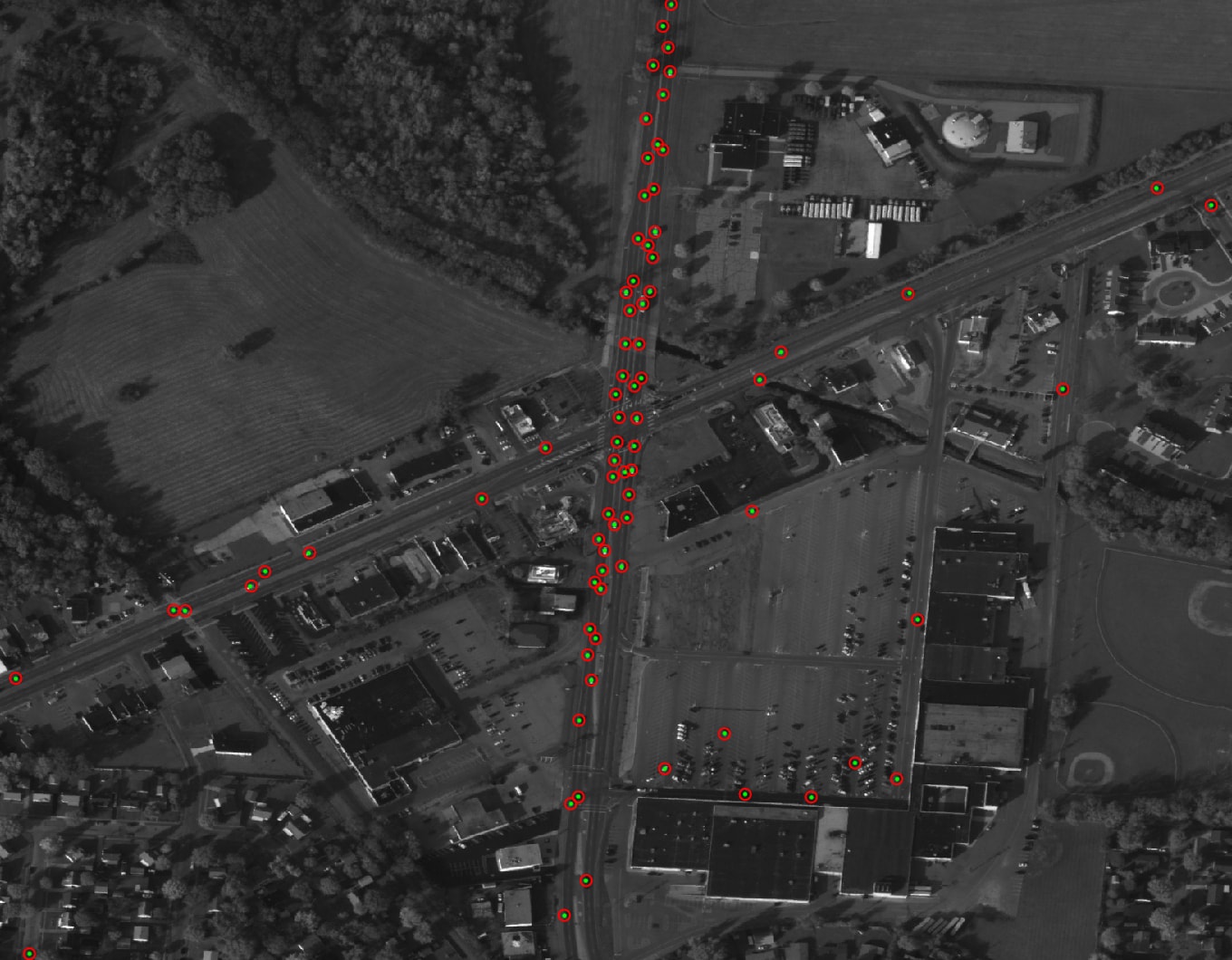}
            \caption{AOI 40 results using 5-frames and the Gaussian heatmap formulation. ClusterNet output shown at left; FoveaNet output and ground-truth shown at right.}
            \label{fig:AOI40ResultAndROC}
        \end{center}
    \end{subfigure}\\[2em]
    \begin{subfigure}[t]{0.95\textwidth}
        \begin{center}
            \includegraphics[width=2.145in]{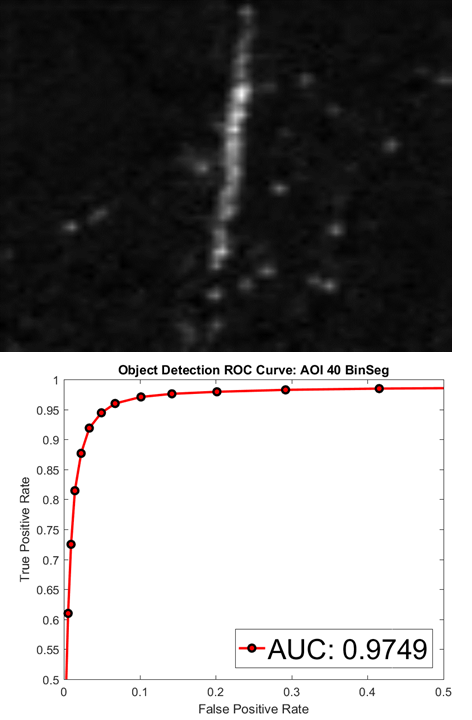}
            \includegraphics[width=4.2in]{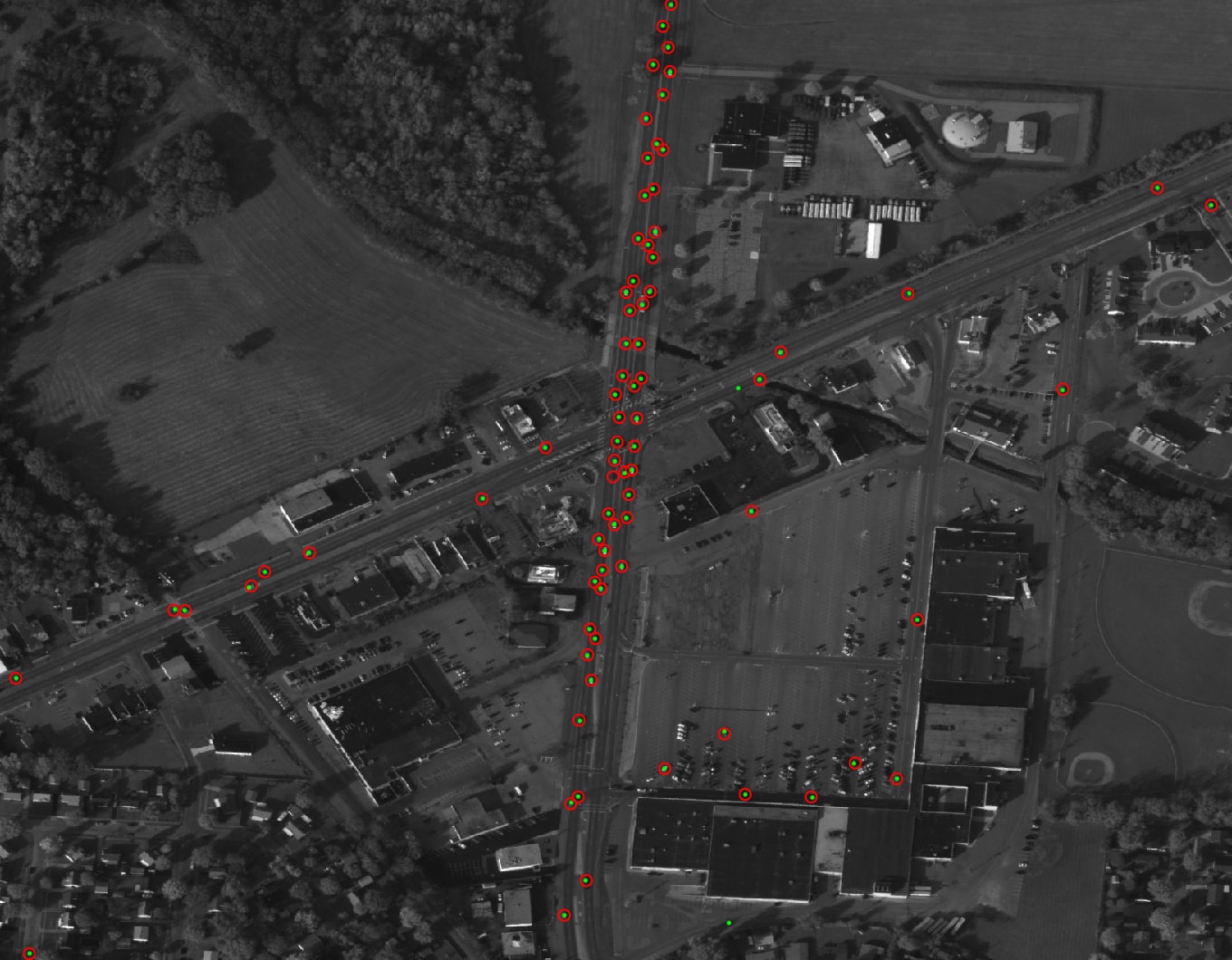}
            \caption{AOI 40 results using 5-frames and the binary segmentation formulation for FoveaNet. ClusterNet output shown at left; FoveaNet output and ground-truth shown at right.}
            \label{fig:AOI40SMResultAndROC}
        \end{center}
    \end{subfigure}
\end{center}
\label{fig:Exemplars5}
\end{figure*}

\addtocounter{figure}{-1}

\begin{figure*}[t]
\begin{center}
    \begin{subfigure}[t]{0.95\textwidth}
        \addtocounter{subfigure}{9}
        \begin{center}
            \includegraphics[width=2.28in]{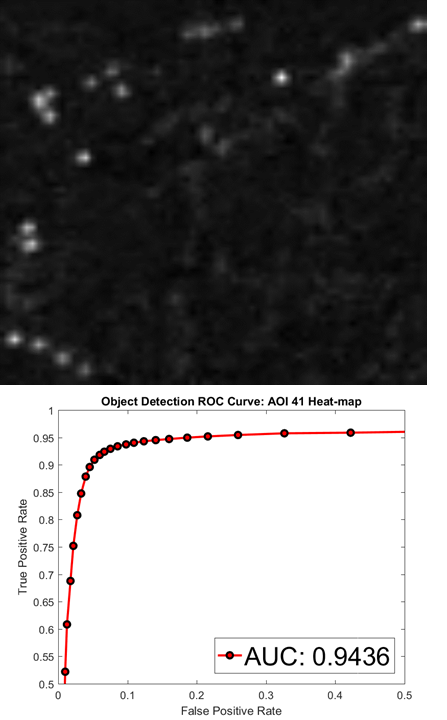}
            \includegraphics[width=4.08in]{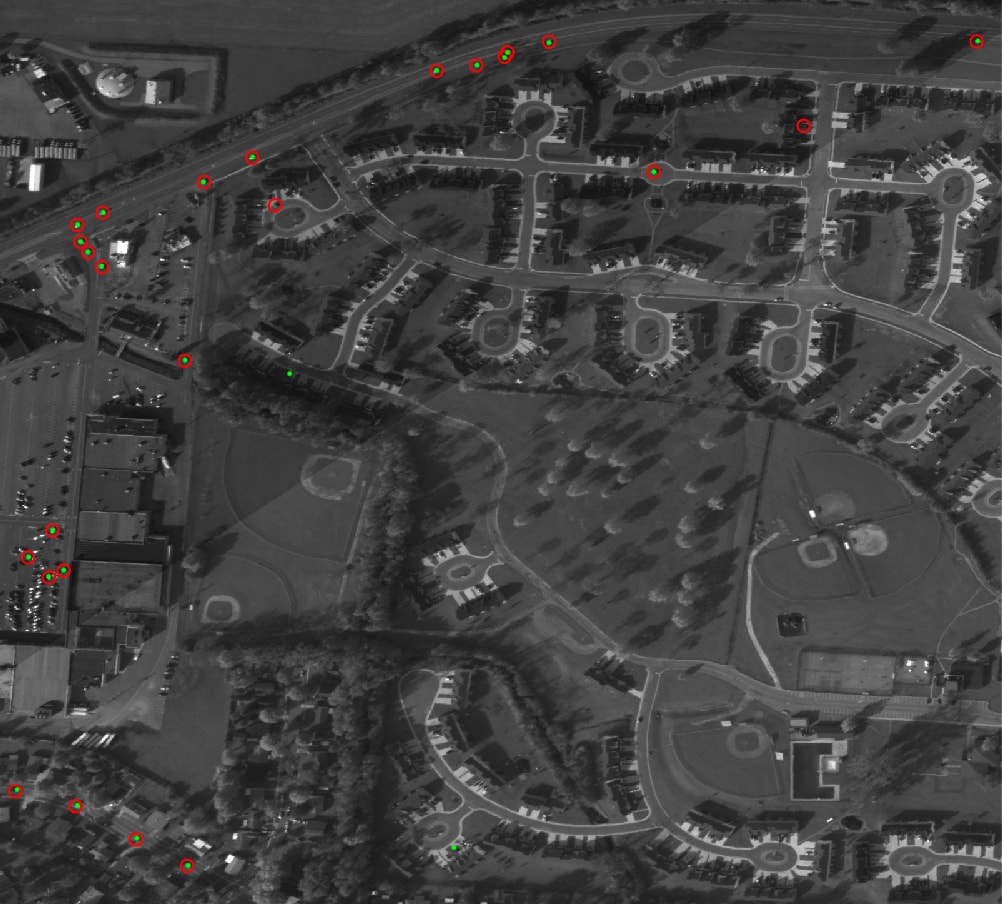}
            \caption{AOI 41 results using 5-frames and the Gaussian heatmap formulation. ClusterNet output shown at left; FoveaNet output and ground-truth shown at right.}
            \label{fig:AOI41ResultAndROC}
        \end{center}
    \end{subfigure}\\[2em]
    \begin{subfigure}[t]{0.95\textwidth}
        \begin{center}
            \includegraphics[width=2.28in]{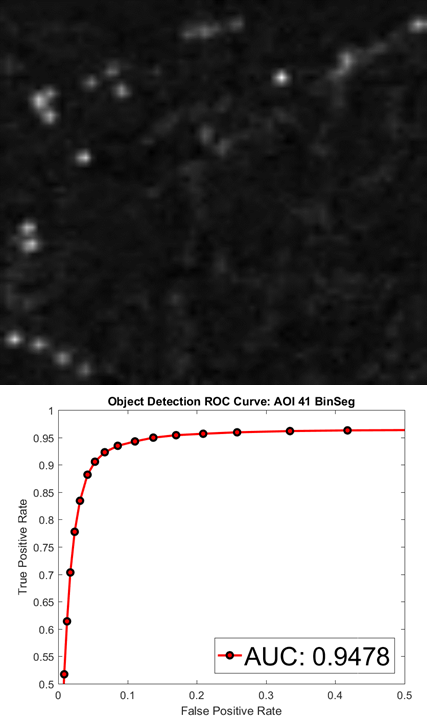}
            \includegraphics[width=4.08in]{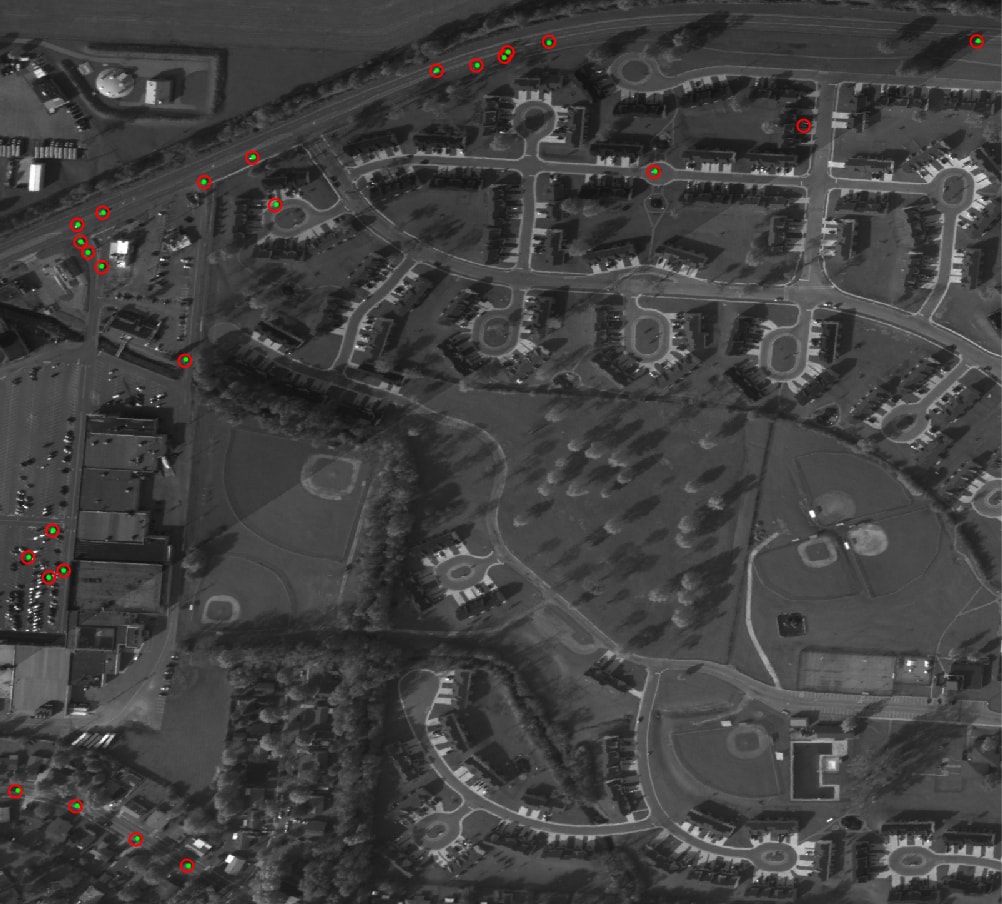}
            \caption{AOI 41 results using 5-frames and the binary segmentation formulation for FoveaNet. ClusterNet output shown at left; FoveaNet output and ground-truth shown at right.}
            \label{fig:AOI41SMResultAndROC}
        \end{center}
    \end{subfigure}
\end{center}
\label{fig:Exemplars6}
\end{figure*}

\addtocounter{figure}{-1}

\begin{figure*}[t]
\begin{center}
    \begin{subfigure}[t]{0.95\textwidth}
        \addtocounter{subfigure}{11}
        \begin{center}
            \includegraphics[width=2.28in]{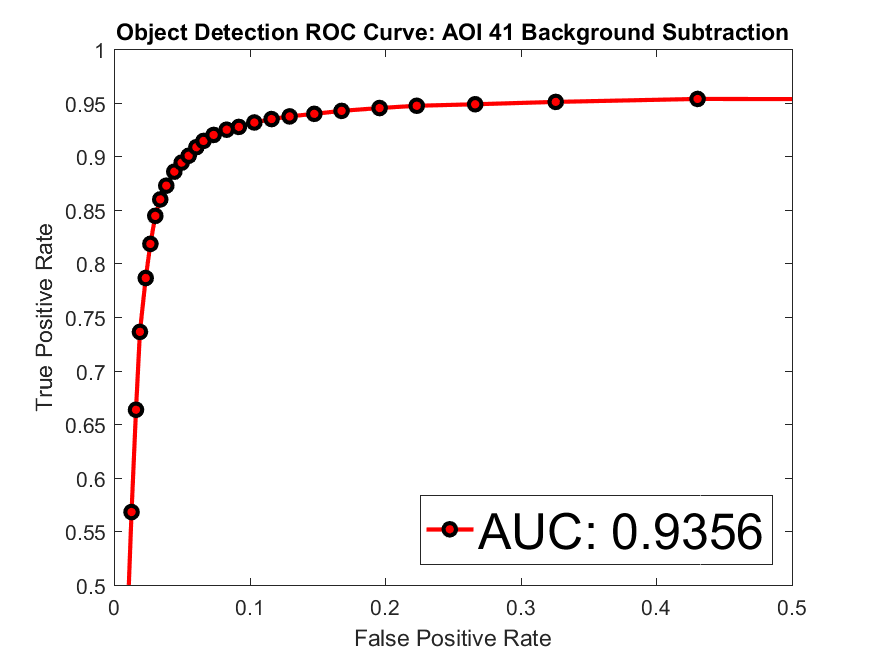}
            \includegraphics[width=4.08in]{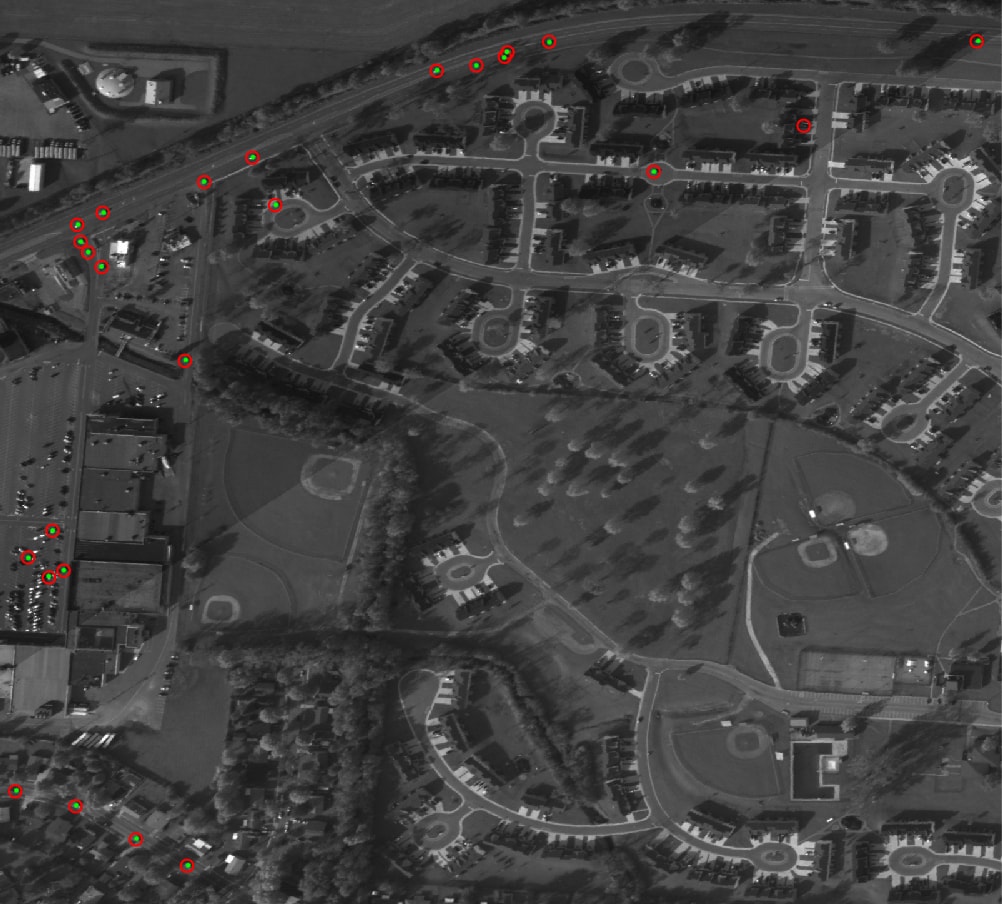}
            \caption{AOI 41 results using the deep learning background subtraction approach.}
            \label{fig:AOI41BSResultAndROC}
        \end{center}
    \end{subfigure}\\[2em]
    \begin{subfigure}[t]{0.95\textwidth}
        \begin{center}
            \includegraphics[width=2.28in]{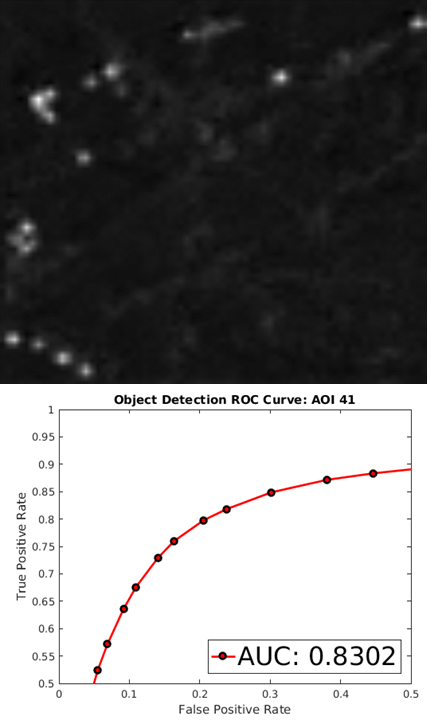}
            \includegraphics[width=4.08in]{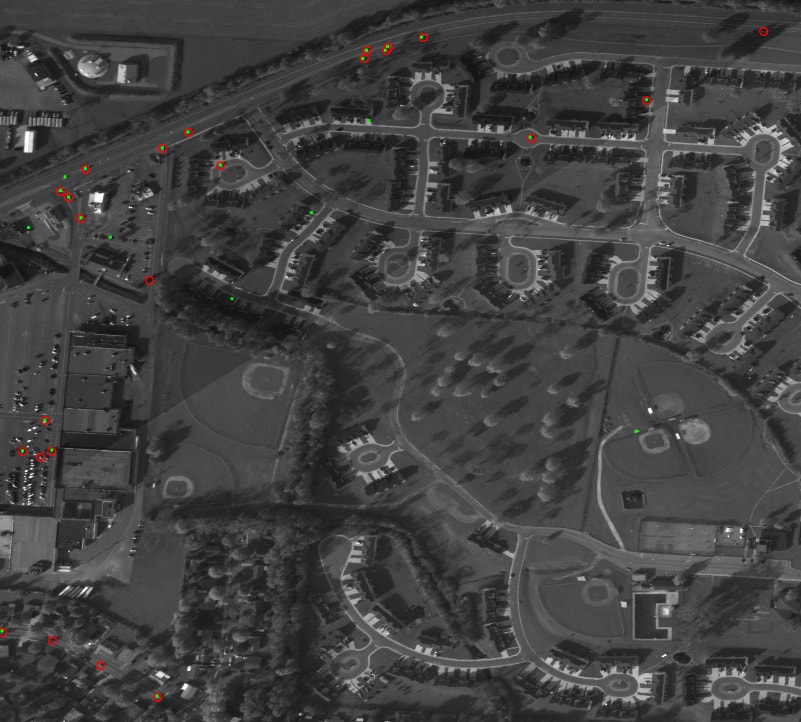}
            \caption{AOI 41 results using 3-frame and the Gaussian heatmap formulation. ClusterNet output shown at left; FoveaNet output and ground-truth shown at right.}
            \label{fig:AOI413FResultAndROC}
        \end{center}
    \end{subfigure}
\end{center}
\label{fig:Exemplars7}
\end{figure*}

\addtocounter{figure}{-1}

\begin{figure*}[t]
\begin{center}
    \begin{subfigure}[t]{0.95\textwidth}
        \addtocounter{subfigure}{13}
        \begin{center}
            \includegraphics[width=2.28in]{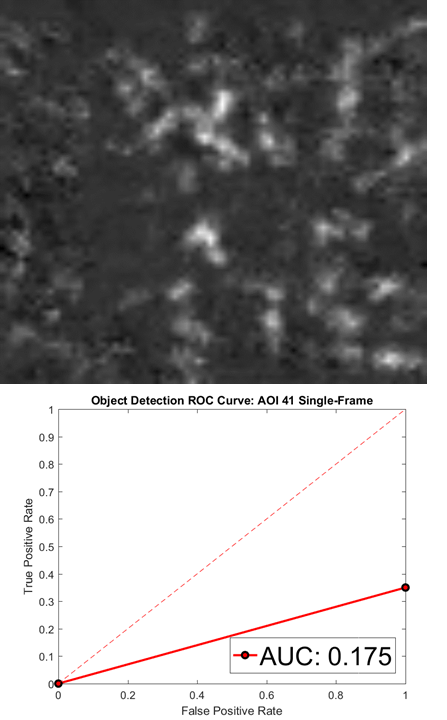}
            \includegraphics[width=4.08in]{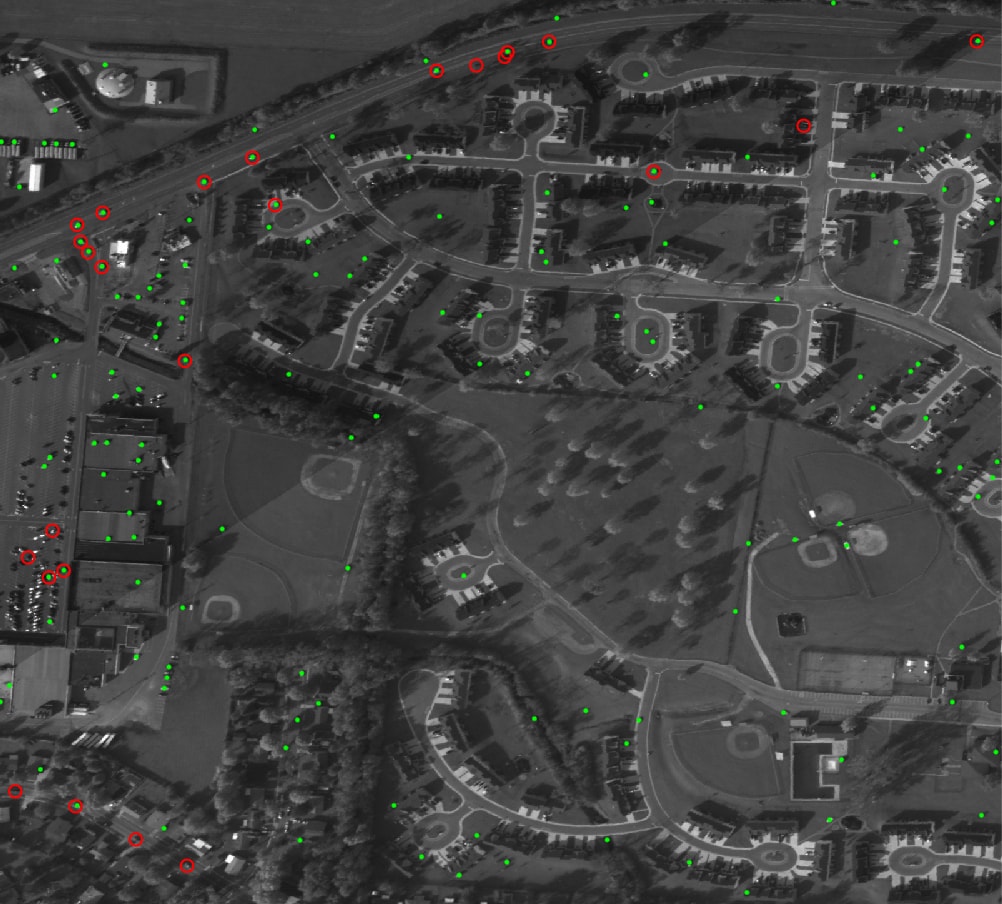}
            \caption{AOI 41 results using 1-frame and the Gaussian heatmap formulation. ClusterNet output shown at left; FoveaNet output and ground-truth shown at right.}
            \label{fig:AOI411FResultAndROC}
        \end{center}
    \end{subfigure}\\[2em]
    \begin{subfigure}[t]{0.95\textwidth}
        \begin{center}
            \includegraphics[width=4.3in]{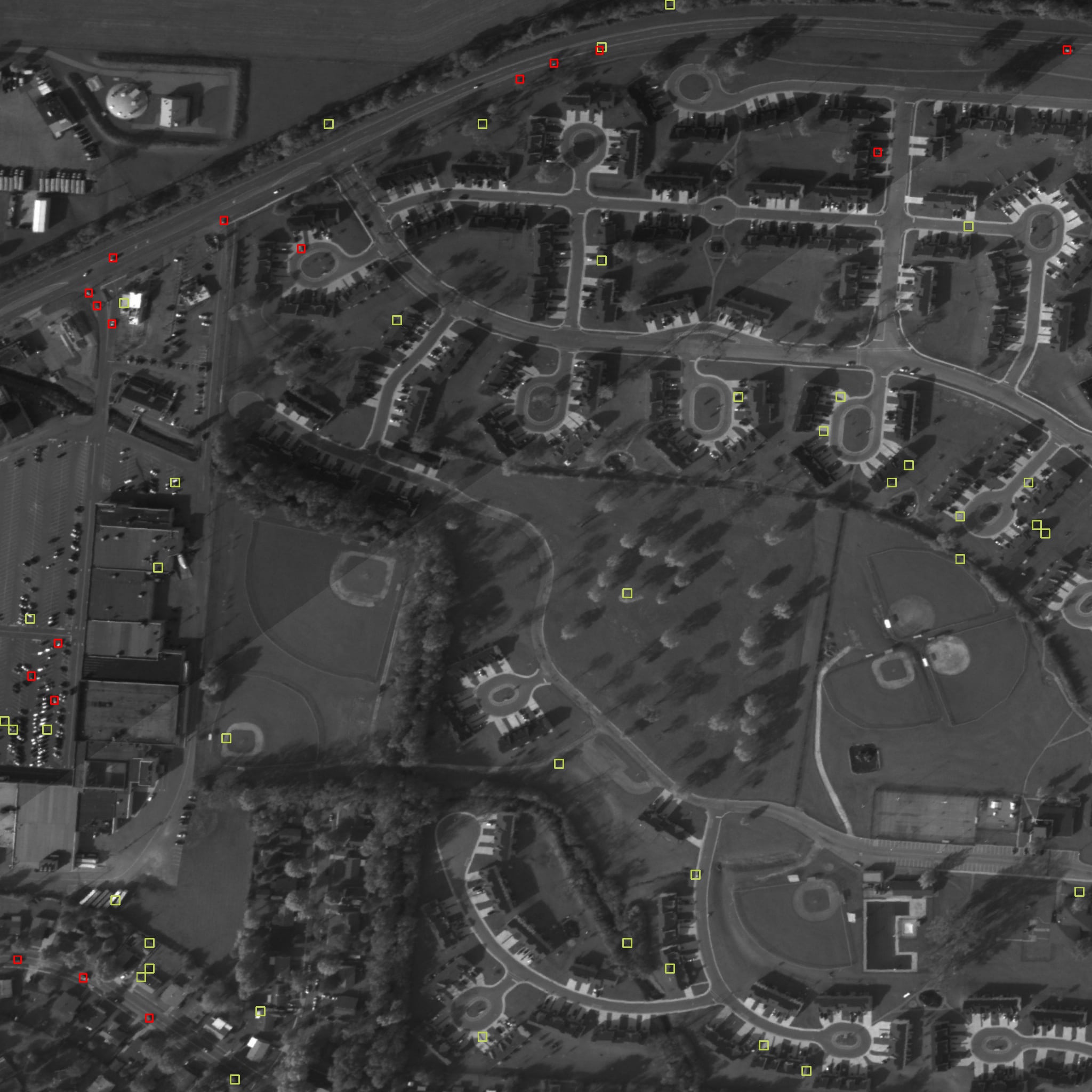}
            \caption{AOI 41 results using Faster R-CNN.}
            \label{fig:AOI41FRCNNResultAndROC}
        \end{center}
    \end{subfigure}
\end{center}
\label{fig:Exemplars8}
\end{figure*}

\end{document}